\documentclass{article} 
\usepackage[preprint]{colm2026_conference}

\usepackage[T1]{fontenc}
\usepackage{microtype}
\usepackage{hyperref}
\usepackage{url}
\usepackage{booktabs}
\usepackage{graphicx}
\usepackage{subfigure}
\usepackage{listings}
\usepackage{xcolor}
\usepackage{colortbl}
\usepackage{tcolorbox}
\usepackage{multirow}
\usepackage{multicol}
\usepackage{amsmath}
\usepackage{amssymb}
\usepackage{mathtools}
\usepackage{amsthm}
\usepackage[capitalize,noabbrev]{cleveref}

\definecolor{darkblue}{rgb}{0, 0, 0.5}
\hypersetup{colorlinks=true, citecolor=darkblue, linkcolor=darkblue, urlcolor=darkblue}

\usepackage{lineno}

\definecolor{gold}{RGB}{255,215,0}
\definecolor{silver}{RGB}{192,192,192}
\definecolor{bronze}{RGB}{205,127,50}

\theoremstyle{plain}

\theoremstyle{definition}

\theoremstyle{remark}

\title{Models for minimalist RAG: B1ade 335M Embedding and 1B Parameter Small Language Models}

\author{Shreyas Subramanian, Mecit Gungor, Vikram Elango \\
Amazon, Seattle, USA \\
\texttt{mecit@amazon.com}
}

\begin{document}

\ifcolmsubmission
\linenumbers
\fi

\maketitle

\begin{abstract}
Language and embedding models used in RAG systems are conventionally assumed to require large-scale pretraining and explicit grounding supervision. We present \textit{B1ade}, an efficient RAG architecture comprising two purpose-built components: a compact embedding model and a purpose-built SLM. B1ade-embed, a 335M parameter retrieval model constructed via parameter-free fusion of five pretrained encoders achieves top MTEB scores among sub-500M models with zero additional training, and B1ade-1B, an SLM trained on low-cost GPUs using Group Relative Policy Optimization (GRPO) on 723M tokens (2.2M examples) of curated context-question pairs with rewards that optimize only answer similarity. Our central finding is emergent attribution: despite receiving no explicit supervision for source citation, B1ade-1B cites retrieved passages in 42.4\% of responses, exceeding the attribution rate of its training distribution by 5.5 percentage points. This demonstrates that grounding behavior can emerge as an accuracy-maximizing strategy under RL training, without explicit reward engineering. On standard QA benchmarks, B1ade-1B achieves 81.82\% on PopQA, 65.8\% on PubMedQA, and 51.09\% on FEVER. In end-to-end RAG evaluation, B1ade-1B achieves an average score of 0.654 across correctness, completeness, coherence, and faithfulness, a 10.8\% improvement over the SFT, while closing the gap with models $1.5\times$ its size. These results show that strategic model composition and reward design suffice for resource-efficient RAG, without large-scale pretraining.
\end{abstract}

\section{Introduction}
\label{sec:intro}

Since its introduction, Retrieval Augmented Generation has emerged as a critical approach for grounding large language model outputs in verifiable sources, particularly for knowledge-intensive tasks in specialized domains~\cite{lewis2020retrieval}. However, deploying effective RAG systems at scale requires addressing two fundamental challenges: (1) efficient retrieval of relevant passages from large corpora, and (2) faithful generation that acknowledges and cites source material. While recent advances in foundation models have improved both retrieval and generation capabilities, state-of-the-art systems typically rely on large models (7B+ parameters) with substantial computational requirements, limiting accessibility for resource-constrained deployments.

A central question for efficient RAG development is whether competitive performance requires extensive pretraining, or whether strategic model composition and targeted optimization can achieve similar capabilities with dramatically reduced compute. We investigate this through two complementary techniques: zero-training model merging for dense retrieval, and direct policy optimization on compact language models for grounded generation. Our approach demonstrates that careful architectural choices and training strategies can match or exceed the per-parameter efficiency of conventional large-scale pretraining.

We introduce \textit{B1ade-embed}, a $335$M parameter embedding model constructed through parameter-free fusion of five pretrained encoders, achieving top performance in sentence similarity among sub-500M models without additional training. This establishes zero-training model merging as a compute-free method for constructing competitive dense retrievers. For generation, we present \textit{B1ade-1B}, a $1$B parameter model trained via Group Relative Policy Optimization (GRPO)~\citep{shao2024deepseekmath} on only $723$M tokens of curated context-question pairs which is orders of magnitude fewer than standard pretraining approaches.

Our primary contribution is the empirical demonstration of emergent attribution behavior in \textit{B1ade-1B} similar to what is described in \cite{wei2022emergent}. Despite training with a reward function that optimizes \textit{exclusively} for answer similarity, with no explicit term for attribution or source citation, the model learns to cite sources in 42.4\% of responses on evaluation datathat is a statistically significant 5.5 percentage point increase over the 36.9\% attribution rate in its training distribution. This emergence validates that grounding behaviors can arise as learned accuracy-maximizing strategies when careful reading and source acknowledgment correlate with correct answers, rather than requiring hand-crafted attribution rewards. The model exhibits strategic application, using specific passage citations at 5.6$\times$ the training rate and applying more attribution to complex multi-hop questions than simple factual queries.

We develop \textit{B1ade-1B} through an iterative process. Our initial model (v1) based on GPT-NeoX with supervised fine-tuning (SFT) achieved suboptimal RAG performance. This motivated a second iteration (v2) combining a Llama-3.2-1B base architecture with GRPO training, substantially improving performance across multiple evaluation dimensions. Empirically, \textit{B1ade-1B} achieves 81.82\% on PopQA, 65.8\% on PubMedQA, and 51.09\% on FEVER, demonstrating that small models trained with minimal compute can achieve competitive performance on standard benchmarks.

\textbf{Contributions:}
\begin{itemize}
\item We introduce zero-training model merging as a compute-free method for constructing competitive dense retrievers, demonstrating that parameter-free fusion of pretrained encoders can achieve top performance without additional training.
\item We show that direct policy optimization on compact language models with simple accuracy rewards induces emergent attribution behavior, with models learning to cite sources 5.5 percentage points more frequently than their training distribution despite no explicit supervision for grounding.
\item We demonstrate that small language models trained via GRPO on $723$M tokens achieve competitive RAG performance, establishing that strategic training can match conventional approaches with orders of magnitude less compute.
\item We provide empirical analysis of the strategy-execution gap in small model attribution, showing that while models learn when to cite sources, factual knowledge limitations remain the primary performance bottleneck.
\end{itemize}

The remainder of this paper is organized as follows. Section~\ref{sec:related} reviews related work in grounded generation, small language models, and reinforcement learning for NLP. Section~\ref{sec:method} describes our training methodology, including the iterative development process and GRPO training approach. We show how the \texttt{simpleCoT} training collection is created, and discuss its unique attribution characteristics. Section~\ref{sec:results} evaluates B1ade across three complementary settings: standard benchmarks, end-to-end RAG, and LLM-as-Judge comparative evaluation, with emphasis on the emergent attribution finding. Section~\ref{sec:discussion} discusses implications, limitations, and evaluation biases identified in our analysis. Section~\ref{sec:conclusion} concludes with a summary of contributions and future directions.

\section{Related Work}
\label{sec:related}

Recent advances in RAG have demonstrated that compact language models can achieve competitive performance when augmented with external knowledge retrieval capabilities \cite{sorstkins2025assessingraghyde1b}. Several compact models in the 1-4 billion parameter range have emerged as viable alternatives to larger architectures, with research establishing 1B parameters as a practical lower bound for effective instruction-following and question-answering capabilities. Notable examples include Pleias-RAG-350M and Pleias-RAG-1B \cite{Langlais2025EvenSR}, which outperform most small language models below 4 billion parameters on standardized RAG benchmarks such as HotPotQA and 2wiki, demonstrating competitive performance with larger 7-8B models including Qwen-2.5-7B and Llama-3.1-8B. These findings suggest that carefully designed compact architectures can achieve strong RAG performance without the resource requirements of their larger counterparts.

To the best of our knowledge, development of competitive embedding models with zero training and model merging is largely unexplored. However, the development of efficient embedding models through knowledge distillation has become a critical component of resource-constrained RAG systems \cite{zhang2025jasperstelladistillationsota}. Multi-stage distillation frameworks enable smaller student embedding models to learn from multiple larger teacher models, addressing the challenge that state-of-the-art embedding models often have numerous parameters and high vector dimensionality that pose deployment challenges in real-world scenarios \cite{chen2025dragdistillingragslms}. Recent work has explored distilling RAG capabilities from large language models to small language models through evidence-based and graph-based knowledge distillation techniques, enabling compact models to replicate retrieval-augmented generation capabilities without extensive fine-tuning \cite{chen2025dragdistillingragslms}. These advances in efficient embedding and knowledge distillation techniques have made RAG more accessible for resource-constrained deployments, aligning with the principle that carefully designed compact components can provide effective solutions for RAG tasks.

\section{Methodology}
\label{sec:method}

We present our approach to resource-efficient RAG through two complementary techniques: zero-training model merging for dense retrieval (\textit{B1ade-embed}, 335M parameters) and direct policy optimization for grounded generation (\textit{B1ade-1B}, 1B parameters). Our methodology demonstrates that strategic model composition and targeted low-cost training can achieve competitive performance with dramatically reduced computational requirements compared to conventional large-scale pretraining.

\subsection{\textit{B1ade-embed}: Zero-Training Model Merging for Dense Retrieval}

\textit{B1ade-embed} is a 335M parameter embedding model created using \emph{Model Stock} model merging, demonstrating that parameter-free fusion of pretrained encoders can achieve competitive performance without additional training. The model comprises the following components using Mergekit\footnote{\url{https://github.com/arcee-ai/mergekit}}~\cite{goddard-etal-2024-arcees}:  (1) \texttt{bert-large-uncased}, (2) \texttt{WhereIsAI/UAE-Large-V1}, (3)  \texttt{BAAI/bge-large-en-v1.5}, (4) \texttt{mixedbread-ai/mxbai-embed-large-v1}, (5)  \texttt{avsolatorio/GIST-large-Embedding-v0}.  Models were chosen based on superior MTEB performance (at the time of writing) despite smaller size and compatible architectures \textit{B1ade-embed} uses the Model Stock algorithm~\cite{mergekit}, which optimizes linear interpolation weights by computing pairwise task vector similarities    
  across models, providing a principled approach to parameter-free model composition. See Appendix~\ref{app:mergekit_details} for detailed       
  merging configuration and algorithm. 

On the \textsc{MTEB} benchmark~\cite{muennighoff2023mteb}, \textit{B1ade-embed} ranks \#4 overall and \#1 in Retrieval. Among models with output dimension $\leq$1024 and fewer than 500M parameters, it achieves \#1 in Sentence Similarity, \#4 in Summarization, Retrieval, and Pair Classification, and \#5 in Reranking. See Appendix~\ref{app:embed_details} for complete MTEB rankings (Table~\ref{tab:mteb_results_detailed}), including a comparison against Stella-400M-v5, the strongest sub-500M retriever which motivates our retriever ablation in Section~\ref{sec:ablation}.

Since its release on the Huggingface Hub, \textit{B1ade-embed} has been independently and externally validated against various unique tasks. For example, in clinical and biomedical tasks, \citet{soffer2024scalable} evaluated 30 models across 2.1M comparisons on clinical notes, synthetic EHRs, MIMIC-IV ICU data, PubMed abstracts, and research papers, identifying \textit{B1ade-embed} as versatile across both clinical and biomedical domains while noting its computational efficiency despite smaller size, particularly strong on short tasks like triage notes and chief complaints where it competed closely with larger models. A related study tested 39 embedding models across 7 medical semantic similarity tasks using Mount Sinai patient data, MIMIC-IV, PubMed texts, and Llama-3-70B synthetic data, finding that smaller models like \textit{B1ade-embed} (335M parameters) performed comparably to larger models on short tasks.

In labor market applications, \citet{qu2026interactive} report that a system incorporating \textit{B1ade-embed} achieved 81\% Positive Predictive Value in a closed-world evaluation against ESCO's occupational hierarchy, demonstrating strong capture of nuanced semantic relationships in labor market terminology.

The syftr (Pareto-Optimal Generative AI) framework~\cite{conway2025syftr} evaluated over $10^{23}$ unique RAG configurations, where \textit{B1ade-embed} consistently appeared in Pareto-optimal solutions, achieving top accuracy on DRDocs (software documentation), CRAG3 (sports), and Infinitebench (long-context retrieval over 100K tokens), while lying on the performance envelope for HotpotQA and Financebench.

\subsection{\textit{B1ade-1B}: Direct Policy Optimization for Grounded Generation}

We develop \textit{B1ade-1B} through an iterative process with two major versions, each addressing specific limitations discovered through evaluation. This iterative development demonstrates the importance of base model selection and training methodology for efficient RAG systems. Table~\ref{tab:model_comparison} summarizes the key differences and improvements between versions.

\begin{table}[t]
\centering
\small
\begin{tabular}{lll}
\toprule
\textbf{Aspect} & \textbf{B1ade v1} & \textbf{B1ade v2} \\
\midrule
\textbf{Base Architecture} & GPT-NeoX-1B & Llama-3.2-1B\\
Pre-training Tokens & 380B & Up to 9T\\
Post-training Tokens & 720M & 720M\\
Epochs & 3 & 1 \\
\textbf{Training Stages} & SFT only &  GRPO only \\
LoRA Rank & 16 & 16 \\
Target Modules & Q, V projection & Q, V projection \\
Trainable Parameters & $\sim$8.4M & $\sim$8.4M \\
Precision & fp32 & bfloat16 \\
\midrule
\multicolumn{3}{l}{\textit{Performance (RAG Evaluation)}} \\
\textbf{Overall Score} & 0.590 & \textbf{0.654 (+10.8\%)} \\
Completeness & 0.435 & \textbf{0.539 (+22.8\%)} \\
Coherence & 0.760 & \textbf{0.856 (+12.6\%)} \\
\bottomrule
\end{tabular}
\caption{Comparison of B1ade v1 and v2 model configurations and performance. B1ade v2 achieves substantial improvements through architectural switch and GRPO training.}
\label{tab:model_comparison}
\end{table}

Our initial model is based on the GPT-NeoX architecture~\cite{andonian2023gpt} with 1 billion parameters, initialized from the pre-trained checkpoint and fine-tuned with Low-Rank Adaptation (LoRA)~\cite{hu2022lora}. While performance on typical text-based benchmarks was superior to other models in the same weight class, evaluation on end-to-end RAG tasks revealed suboptimal performance as shown in our results, particularly in completeness and coherence dimensions, motivating architectural and training improvements.

Based on these limitations on end-to-end RAG tasks, we developed a second version of \textit{B1ade} with two key changes: (1) architecture switch to Llama-3.2-1B~\cite{llama3} base, and (2) training enhancement directly using GRPO with no SFT checkpoint, since Llama-3.2-1B provides a useful and competitive base, which we include in our evaluations. 

\subsection{Training Data: simpleCoT Collection}
\label{sec:dataset}

We construct \texttt{simpleCoT}, a large-scale dataset of 2,214,941 examples (723M tokens) aggregated from seven sources spanning multi-hop QA, instruction following, mathematical reasoning, and commonsense understanding. All splits are unified into a standardized schema and shuffled with fixed seed 42, yielding 1,771,953 training and 442,988 test examples. Full dataset composition and processing details are provided in Appendix~\ref{app:dataset_details}. The dataset is publicly available.\footnote{\url{https://huggingface.co/datasets/w601sxs/simpleCoT}}

\subsection{Training}

We directly perform GRPO~\citep{shao2024deepseekmath} on top of the Llama-3.2-1B base using a standard ROUGE-L reward on answer similarity, with no explicit attribution or grounding term. We train for 3 epochs with learning rate $5 \times 10^{-6}$ (AdamW), group size $k{=}4$, batch size 8 questions (32 responses), bfloat16 precision, on a single NVIDIA A10G (24GB) for approximately 12 hours.

Critically, our reward function optimizes only for answer similarity using ROUGE-L score, with no explicit term for attribution or grounding behavior. This design choice is intentional: we investigate whether grounding behavior can emerge as a learned accuracy-maximizing strategy when careful reading and source acknowledgment correlate with correct answers, without requiring explicit supervision or hand-crafted attribution rewards. This minimal reward design enables us to isolate and study the emergence phenomenon.

\section{Empirical Evaluation}
\label{sec:results}

We evaluate our approach across two complementary settings: (1) standard context-based QA benchmarks to measure core reasoning capabilities, and (2) end-to-end RAG to assess real-world applicability with retrieval systems using LLM-as-Judge scoring. Our evaluation focuses on validating the emergent attribution phenomenon and characterizing the performance-efficiency tradeoffs of our training paradigm.

As described in Section~\ref{sec:method}, we develop two versions of \textit{B1ade-1B}: v1 (GPT-NeoX base, SFT only) and v2 (Llama-3.2-1B base,  GRPO-only). Standard benchmarks and LLM-as-Judge evaluations use v2, while RAG evaluation compares both versions to demonstrate the improvement from architectural and training changes. Due to the poor performance of the base v1 model on end-to-end RAG tasks, we do not use the same for subsequent GRPO training, or for ablation experiments such as SFT-only vs GRPO-only. We focus on showing that simple reward functions (like rouge score) can still provide meaningful improvements in tasks like RAG. Next we discuss standard QA benchmarks and our end-to-end RAG benchmarks involving both B1ade-embedding and B1ade SLM.

\subsection{Standard Context-Based QA Benchmarks}
\label{sec:standard_benchmarks}

We evaluate \textit{B1ade-1B} on standard question-answering benchmarks (PopQA, TriviaQA, Natural Questions, WikiMultihopQA, FEVER, ARC-Challenge, PubMedQA) using Exact Match accuracy. See Appendix~\ref{app:data_scale_benchmarks} for the complete evaluation protocol.

\begin{table*}[h!]
\centering
\caption{Standard context-based QA benchmark results (Exact Match \%). \textit{B1ade-1B} achieves the highest PopQA score (81.82\%) and strong performance on PubMedQA (``PMQA'') (65.80\%) and FEVER (51.09\%) among 1B models. $0^*$~indicates instruction-following failures.}
\label{tab:standard_benchmarks}
\vskip 0.1in
\resizebox{\textwidth}{!}{%
\begin{tabular}{@{}lccccccccc@{}}
\toprule
\textbf{Model} & \textbf{Size} & \textbf{PopQA} & \textbf{TQA} & \textbf{NQ} & \textbf{ASQA} & \textbf{FEVER} & \textbf{ARC} & \textbf{PMQA} & \textbf{WikiMQA} \\
\midrule
TinyLlama \cite{pap1} & 1.1B & 64.32 & \textbf{60.00} & 44.10 & 0.48 & $0^*$ & 23.46 & $0^*$ & 0.23 \\
Llama-3.1B \cite{llama3} & 1.0B & 48.55 & 36.65 & 35.67 & \textbf{40.21} & 23.27 & 49.91 & 16.8 & \textbf{12.79} \\
Phi-1.5 \cite{pap3} & 1.2B & 68.30 & 50.34 & 0 & 0.14 & $0^*$ & 51.96 & 34.4 & 0.04 \\
OPT \cite{pap4} & 1.3B & 70.51 & 43.75 & 47.80 & 2.75 & $0^*$ & 22.95 & $0^*$ & 0 \\
Pythia \cite{pap5} & 1.0B & 73.44 & 39.77 & \textbf{66.90} & 0.01 & 8.17 & 22.78 & 18.2 & 0.06 \\
Qwen-2.5 \cite{pap6} & 1.5B & 74.77 & 41.10 & 51.36 & 37.04 & 23.98 & \textbf{71.93} & 57.8 & 11.59 \\
\midrule
\textbf{B1ade-1B} & \textbf{1.0B} & \textbf{81.82} & 33.11 & 48.32 & 16.56 & \textbf{51.09} & 22.44 & \textbf{65.80} & 7.56 \\
\bottomrule
\end{tabular}%
}
\end{table*}

\begin{figure}[h]
    \centering
    \includegraphics[width=0.8\linewidth]{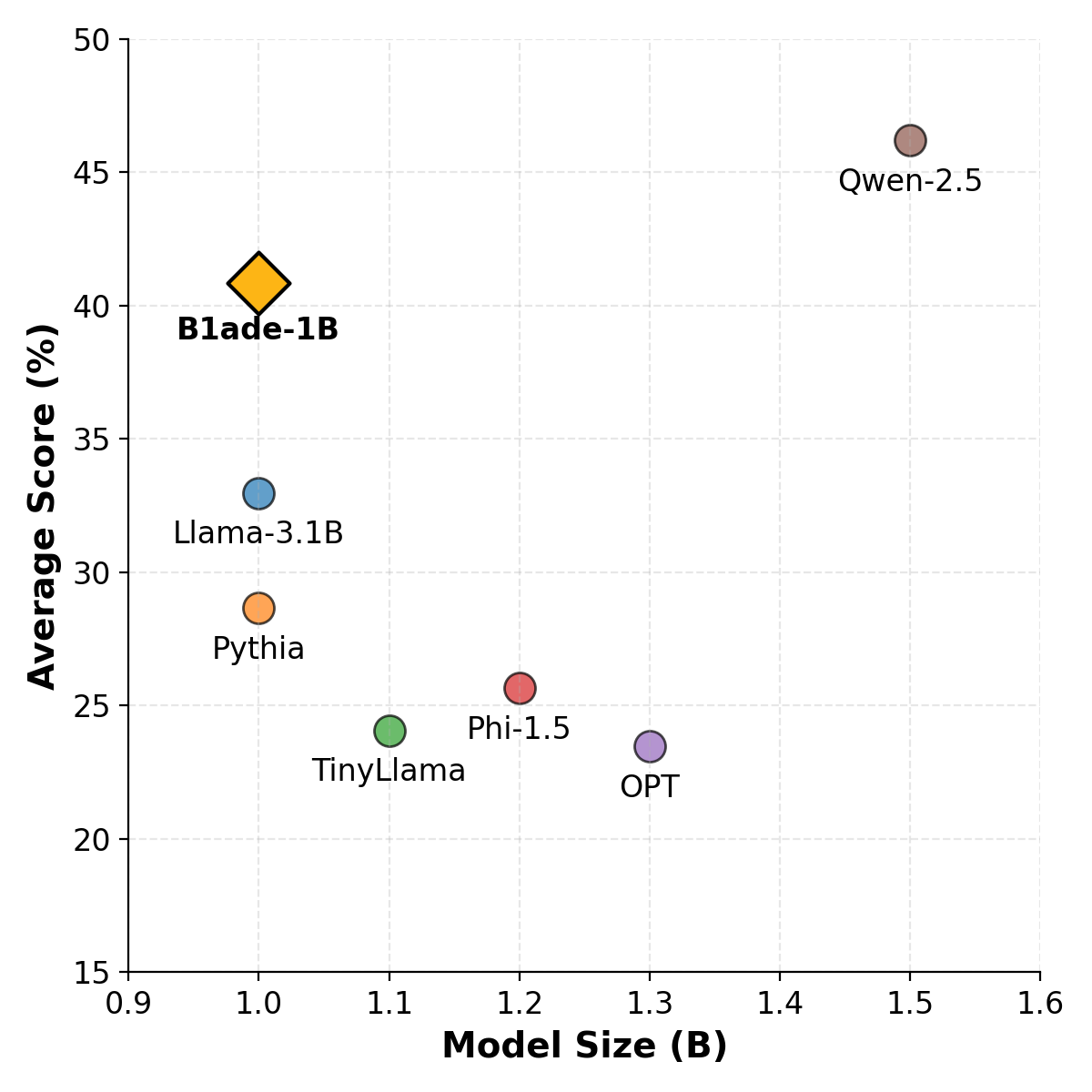}
    \caption{B1ade performance on standard benchmarks}
    \label{fig:benchmarks2}
\end{figure}

\textit{B1ade-1B} achieves 81.82\% on PopQA (surpassing Qwen-2.5's 74.77\%), 65.80\% on PubMedQA, and 51.09\% on FEVER. These results show that training on context-based tasks with reasoning chains enables transfer across domains. Our model surpasses Qwen-2.5 on three tasks despite Qwen having 50\% more parameters, indicating that training on curated data compensates for reduced capacity. The higher average scores relative to other 1B models  result from strong performance on key tasks and avoidance of catastrophic failures observed in other sub-2B models (Also see Fig.~\ref{fig:benchmarks2} for a visual). For other models, zero or near-zero scores result from instruction-following failures and strict exact-match evaluations (detailed analysis in Appendix~\ref{sec:zeroscores}). Among the models we tested, only Llama-1B and \textit{B1ade-1B} achieve consistent performance without catastrophic failures. Other models exhibit context repetition, option echoing, and format inconsistencies, revealing limitations in deploying sub-2B models for structured RAG without constrained decoding (detailed in Appendix~\ref{app:examples}). 

\subsection{End-to-End RAG Evaluation}
\label{sec:rag_eval}

We evaluate \textit{B1ade-1B} in realistic RAG settings using \textit{B1ade-embed} and stella-400M for retrieval, with Claude Sonnet 4 as LLM-as-Judge on correctness, completeness, coherence, and faithfulness. We use LLM-as-Judge-based model evaluation jobs on Amazon Bedrock; this evaluation motivated our v1 to v2 improvements.\footnote{Built-in model evaluation metrics: \url{https://docs.aws.amazon.com/bedrock/latest/userguide/model-evaluation-metrics.html}}

\begin{table*}[h!]
\centering
\caption{End-to-end RAG evaluation results using B1ade-embed (Claude Sonnet 4 as judge). Scores range from 0--1 (higher is better). \textbf{Bold} denotes global best per metric. See Section~\ref{sec:ablation} for retriever choice and training data scale ablations.}
\label{tab:rag_results}
\vskip 0.1in
\begin{tabular}{p{2.0cm}p{2cm}p{0.5cm}p{0.8cm}p{0.8cm}p{0.8cm}p{0.8cm}p{0.8cm}}
\toprule
\textbf{LLM} & \textbf{Embedding} & \textbf{K} & \textbf{Corr.} & \textbf{Comp.} & \textbf{Cohe.} & \textbf{Faith.} & \textbf{Avg.} \\
\midrule
\multicolumn{8}{@{}l}{\textit{Baseline Models (no fine-tuning)}} \\
Llama-1B & \textit{B1ade-embed}  & 5 & 0.636 & 0.549 & 0.911 & 0.639 & \textbf{0.684} \\
Qwen-1.5B & \textit{B1ade-embed}  & 5 & \textbf{0.679} & 0.527 & 0.829 & 0.680 & 0.679 \\
\midrule
\multicolumn{8}{@{}l}{\textit{B1ade v1 (GPT-NeoX Base, SFT Only)}} \\
B1ade v1 & \textit{B1ade-embed}  & 5 & 0.521 & 0.435 & 0.760 & 0.644 & 0.590 \\
\midrule
\multicolumn{8}{@{}l}{\textit{B1ade v2 (Llama-3.2-1B Base, GRPO)}} \\
\textit{B1ade v2} & \textit{B1ade-embed} & 5 & 0.541 & 0.539 & 0.856 & \textbf{0.681} & \textit{0.654} \\
\bottomrule
\end{tabular}%
\end{table*}

B1ade v1 (GPT-NeoX + SFT) scores 0.590 average, with low completeness (0.435) and coherence (0.760). Switching to Llama-3.2-1B and adding GRPO (v2) improves to 0.654 (+10.8\%), with largest gains in completeness (+22.8\%) and coherence (+12.6\%). B1ade v2 approaches Qwen-1.5B (0.679) with 33\% fewer parameters but trails in correctness (0.541 vs.\ 0.679). While Llama-1B base achieves the highest average score (0.684) driven by coherence (0.911), B1ade v2 leads all models on faithfulness (0.681), the metric most directly reflective of RAG grounding quality, demonstrating that GRPO training successfully optimizes for source-grounded generation despite using 33\% fewer parameters than Qwen-1.5B. Also see note in Appendix \ref{sec:biasllm} on length bias from judge model impacting Llama 1B scores.

\subsection{Attribution Emergence Analysis}
\label{sec:llmaaj}

We conduct head-to-head comparisons using LLM-as-Judge on 6,345 QA pairs from RAGEval~\cite{rageval}, comparing \textit{B1ade} v2, Llama-3.2-1B (base), and Qwen-1.5B. Claude Sonnet 4 ranks outputs based on context fidelity, attribution quality, completeness, and factual accuracy (see Appendix~\ref{sec:cmpletetemplate} for the complete template). This evaluation enables direct measurement of the emergent attribution phenomenon.

\begin{table}[t]
\centering
\caption{LLM-as-Judge evaluation results ($n = 6{,}345$). \textit{Top:} Primary rankings showing B1ade v2 achieves 32.2\% first-place rate with our training approach (Llama-3.2-1B base with GRPO). \textit{Bottom:} Attribution emergence---B1ade-1B exceeds the training distribution by 5.5 percentage points, representing the main novel contribution of this work.}
\label{tab:llmaaj_results}
\vskip 0.1in
\small
\begin{tabular}{@{}lcccc@{}}
\toprule
\textbf{Rank} & \textbf{Model} & \textbf{First Place} & \textbf{Avg Rank} & \textbf{Params} \\
\midrule
1st & Qwen-1.5B & \textbf{42.2\%} (2,678) & 1.77 & 1.5B \\
2nd & B1ade v2 & \textbf{32.2\%} (2,044) & 2.11 & 1B \\
3rd & Llama-3.2-1B & \textbf{29.1\%} (1,849) & 2.08 & 1B \\
\midrule
\multicolumn{2}{@{}l}{Training Data Attribution} & \multicolumn{2}{c}{36.9\%} & \\
\multicolumn{2}{@{}l}{B1ade Eval Attribution} & \multicolumn{2}{c}{\textbf{42.4\%}} & \\
\bottomrule
\end{tabular}
\end{table}

\textit{B1ade-1B} achieves 32.2\% first-place rate vs.\ 29.1\% for base Llama-1B (+3.1pp) and reaches 76\% of Qwen-1.5B's performance (42.2\%) with 67\% of its parameters. The key finding is that \textit{B1ade-1B} uses attribution at 42.4\%, exceeding the 36.9\% rate found naturally in the training distribution. This shows grounding emerges as an accuracy-maximizing strategy, not through memorization.

With 6,345 evaluations, the 5.5pp increase (2,689 observed vs.\ 2,341 expected) is significant ($p < 0.001$, binomial test). If the model merely memorized training patterns, we would expect 36.9\% attribution and uniform application. Instead we observe: (1) 42.4\% attribution (+5.5pp above training), (2) specific citations at 2.8\% (5.6$\times$ training rate), and (3) higher attribution on multi-hop vs.\ factual questions. 

\subsection{50K Subset Curation Methodology}
\label{sec:subset_curation}

To enable fair comparison of GRPO's effectiveness across different data scales, we created a clean 50K subset from the 2.2M-example simpleCoT dataset using a four-step pipeline:

\begin{enumerate}
\item \textbf{Normalization:} Extract raw question and answer text from dataset-specific XML-like scaffolding (e.g., \texttt{question: <...>}, \texttt{answer: <...>}). Drop malformed examples.

\item \textbf{Length filtering:} Retain only examples where the question contains 20--300 tokens and the answer contains 10--200 tokens. This eliminates trivial one-word answers and excessively long documents while maintaining diversity in question complexity.

\item \textbf{Deduplication:} Remove exact duplicate questions by hashing the first 80 characters of cleaned question text, preserving the first occurrence.

\item \textbf{Stratified sampling:} Partition questions into four equal-sized difficulty bins based on token length quartiles (easy → hard), then sample uniformly from each bin to ensure balanced coverage of question complexity. This prevents the subset from being skewed toward either trivial or overly complex questions.
\end{enumerate}

This curation process reduces the 2.2M raw examples to approximately 500K candidate examples after filtering, from which we stratified-sample 50K for the scale ablation. The resulting subset is deterministic (seeded), publicly available on Hugging Face Hub\footnote{\url{https://huggingface.co/datasets/w601sxs/simplecot_subset_50k}}, and serves as the training corpus for the smaller model experiments in Section~\ref{sec:ablation}. Note: 50K experiments train for 1 epoch, while full 2.2M experiments train for 3 epochs.

\subsection{Ablation Studies and Additional Analyses}
\label{sec:ablation}

We present three complementary analyses examining training data scale, retriever choice, and inference optimization.

\textbf{Training data scale: 50K vs.\ 2.2M.} To assess how much training data is necessary to realize GRPO's benefits, we train Qwen-0.5B and Llama-1B on only 50K examples from the simpleCoT dataset. Table~\ref{tab:data_scale_ablation} (RAG) and Appendix Table~\ref{tab:data_scale_benchmarks} (standard benchmarks) show that standard benchmark performance is substantially lower than \textit{B1ade-1B} trained on the full 2.2M examples, confirming that factual knowledge acquisition requires large-scale training.

GRPO's effects differ across data scales. On 50K, GRPO improves all Llama-1B metrics except faithfulness: correctness +0.047 (+12.9\%), coherence +0.066 (+18.6\%), completeness +0.003 (+0.8\%). On 2.2M, B1ade v2 shows the opposite: only faithfulness improves (+0.042, +6.6\%), while correctness declines -0.095, coherence -0.055, completeness -0.010 relative to base Llama-1B. With limited data, GRPO improves general quality. With abundant data, GRPO optimizes exclusively for grounding at the expense of other metrics. This indicates GRPO's reward does not scale uniformly. GRPO reduces performance on format-constrained tasks (FEVER: -13.1pp for Qwen-0.5B), consistent with the reward not penalizing format deviations. These results suggest decoupling between knowledge acquisition (scale-dependent) and grounding behavior (learnable from limited data).

\begin{table}[h]
\centering
\caption{Training data scale ablation: Impact of data quantity on GRPO effectiveness (RAG evaluation with B1ade-embed, scores 0--1). Bottom row shows Llama-1B trained on full 2.2M dataset with GRPO (B1ade v2) for comparison.}
\label{tab:data_scale_ablation}
\vskip 0.05in
\begin{tabular}{lccccc}
\toprule
\textbf{Model} & \textbf{Data} & \textbf{Corr.} & \textbf{Comp.} & \textbf{Faith.} & \textbf{Avg} \\
\midrule
Qwen-0.5B & 50K & 0.321 & 0.356 & 0.393 & 0.345 \\
Qwen-0.5B GRPO & 50K & 0.340 & 0.369 & 0.425 & 0.368 \\
\midrule
Llama-1B & 50K & 0.364 & 0.383 & 0.430 & 0.383 \\
Llama-1B GRPO & 50K & 0.411 & 0.386 & 0.421 & 0.407 \\
\midrule
\textit{B1ade v2 (Llama-1B)} & \textit{2.2M} & \textit{0.541} & \textit{0.539} & \textit{0.681} & \textit{0.654} \\
\bottomrule
\end{tabular}
\end{table}

\textbf{Retriever choice (B1ade-embed vs.\ Stella-400M).} Table~\ref{tab:retriever_ablation} shows minimal impact from retriever choice. B1ade v2 scores 0.654 with B1ade-embed vs.\ 0.652 with Stella-400M (+0.002 difference); B1ade v1 shows 0.590 vs.\ 0.589. Generation quality is bottlenecked by the language model, not retrieval at $K{=}5$. B1ade-embed is a compute-free alternative to Stella-400M without quality loss.

\begin{table}[h]
\centering
\caption{Retriever ablation: B1ade-embed vs.\ Stella-400M on B1ade v1 and v2 (scores range 0--1, higher is better).}
\label{tab:retriever_ablation}
\vskip 0.05in
\begin{tabular}{lcccc}
\toprule
\textbf{Model} & \textbf{Embedding} & \textbf{Corr.} & \textbf{Comp.} & \textbf{Avg} \\
\midrule
B1ade v1 & B1ade-embed & 0.521 & 0.435 & 0.590 \\
B1ade v1 & Stella-400M & 0.526 & 0.437 & 0.589 \\
\midrule
B1ade v2 & B1ade-embed & 0.541 & 0.539 & 0.654 \\
B1ade v2 & Stella-400M & 0.545 & 0.542 & 0.652 \\
\bottomrule
\end{tabular}
\end{table}

\textbf{Retrieval scope (K=5 vs K=10).} We test retrieval scope to assess sensitivity to the number of retrieved documents. Table~\ref{tab:k_ablation} compares B1ade v2 with K=5 (0.654 avg) and K=10 (0.466 avg). Increasing K from 5 to 10 documents provides minimal benefit and  degrades performance, consistent with the observation that language model capacity is the bottleneck. We use K=5 for all reported results.

\begin{table}[h]
\centering
\caption{Retrieval scope ablation: B1ade v2 with K=5 vs K=10 (B1ade-embed, scores 0--1).}
\label{tab:k_ablation}
\vskip 0.05in
\begin{tabular}{lccccc}
\toprule
\textbf{Model} & \textbf{K} & \textbf{Corr.} & \textbf{Comp.} & \textbf{Cohe.} & \textbf{Avg} \\
\midrule
B1ade v2 & 5 & 0.541 & 0.539 & 0.856 & 0.654 \\
B1ade v2 & 10 & 0.431 & 0.347 & 0.566 & 0.466 \\
\bottomrule
\end{tabular}
\end{table}

\textbf{Inference optimization: speculative decoding.} We test self-speculative decoding for inference throughput. DoLa decoding reduces PubMedQA accuracy from 65.8\% to 63.8\%, while ngram lookup improves it to 68.2\%. On FEVER, both methods degrade accuracy by 25\% due to repeated token artifacts. Speculative decoding disrupts the attribution patterns learned during GRPO training.

\section{Discussion and Limitations}
\label{sec:discussion}

We organize this discussion around key themes: the emergence and validation of attribution as a learned strategy, the strategy-execution gap revealed by our evaluations, implications for efficient RAG development, and limitations with future directions.

B1ade-1B exhibits attribution at 42.4\%, exceeding training distribution (36.9\%) by 5.5pp. This validates that grounding emerges as an accuracy-maximizing strategy. However, the base Llama-3.2-1B may have pre-existing attribution capabilities from distillation or prior training. We cannot definitively separate GRPO training effects from base model capabilities.

Evidence of learned behavior beyond base capabilities: (1) Specific passage citations (e.g., ``Passage 1 states...'') appear at 2.8\%, 5.6$\times$ the training rate of 0.5\%, with insufficient training examples to explain via memorization. (2) Attribution rate is higher on multi-hop questions than factual queries, not uniform across types. (3) Direct quotes appear in 9.8\% of responses, with 494 accurate (38.8\%), showing text extraction beyond templates. (4) The model correctly distinguishes when context supports vs.\ does not support claims (e.g., recognizing ``passage does not explicitly mention emergence, only predominance''), showing epistemic calibration.

Regarding training limitations, the simpleCoT dataset containing 2.2M examples exhibits source imbalance, with Kaist contributing 76.6\% of training data (Table~\ref{tab:source_distribution_app}). This may limit generalization to domains underrepresented in training, such as creative writing or low-resource languages. Regarding evaluation limitations, despite explicit instructions to avoid length bias, we observe systematic length-rank correlation (Table~\ref{tab:length_bias}), with Llama's longer responses penalized (correlation of +0.264) and Qwen's brevity rewarded (correlation of -0.112). The ``attribution paradox,'' where B1ade-1B loses to Qwen despite providing identical answers with explicit attribution (Appendix Cases 1--3), suggests the judge may favor conciseness over source citation, contradicting stated preferences. All LLM-as-Judge evaluations use Claude Sonnet 4 as a single judge model. Different judge models such as GPT-4 or Llama-3-70B-Instruct may exhibit different biases, potentially altering rankings, and multi-judge consensus would strengthen validity. Our standard benchmarks in Section~\ref{sec:standard_benchmarks} focus on factual QA and reasoning. We do not evaluate creative generation, dialogue, or code generation, which are domains where attribution behavior may manifest differently.

\section{Conclusion}
\label{sec:conclusion}

We introduce a training paradigm for resource-efficient RAG using zero-training model merging for retrieval and policy optimization on compact language models. Strategic architectural choices and training achieve competitive performance with reduced computational requirements. Through this training paradigm We presented B1ade, a resource-efficient RAG system with two contributions: B1ade-embed, a competitive 335M retriever built via zero-training model merging that matches Stella-400M in end-to-end RAG (0.654 vs.\ 0.652) without any retriever training; and B1ade-1B, a 1B SLM trained with GRPO on 723M tokens that achieves 81.82\% PopQA, 65.8\% PubMedQA, and 0.654 average RAG score (LLaaJ) that approaches Qwen-1.5B with 500M fewer parameters. An emergent property of GRPO training is that attribution behavior arises through accuracy-only optimization: training with ROUGE rewards alone produces a 42.4\% citation rate, exceeding the training distribution (36.9\%) by 5.5\%. Together, these results show that competitive RAG components can be built without large-scale specialized training, with grounding behavior emerging as a byproduct of reward optimization rather than explicit supervision.

Future work should investigate whether attribution emergence generalizes across model scales, training curricula, and reward formulations. Key questions include: (1) Does this phenomenon persist in larger models, or is it specific to the capacity constraints of 1B-scale systems? (2) Can alternative reward formulations (e.g., token-level rewards, contrastive objectives) induce stronger or more reliable grounding behaviors? (3) What training data characteristics (attribution rate, citation specificity, reasoning chain length) most strongly influence emergence? Additionally, methods to close the strategy-execution gap without sacrificing parameter efficiency warrant investigation, potentially through knowledge distillation or hybrid architectures that separate reasoning from knowledge storage.

\bibliography{example_paper}

@article{pap1,
  title={Tinyllama: An open-source small language model},
  author={Zhang, Peiyuan and Zeng, Guangtao and Wang, Tianduo and Lu, Wei},
  journal={arXiv preprint arXiv:2401.02385},
  year={2024}
}

@article{pap6,
      title={Qwen2 Technical Report}, 
      author={An Yang and Baosong Yang and Binyuan Hui and Bo Zheng and Bowen Yu and Chang Zhou and Chengpeng Li and Chengyuan Li and Dayiheng Liu and Fei Huang and Guanting Dong and Haoran Wei and Huan Lin and Jialong Tang and Jialin Wang and Jian Yang and Jianhong Tu and Jianwei Zhang and Jianxin Ma and Jin Xu and Jingren Zhou and Jinze Bai and Jinzheng He and Junyang Lin and Kai Dang and Keming Lu and Keqin Chen and Kexin Yang and Mei Li and Mingfeng Xue and Na Ni and Pei Zhang and Peng Wang and Ru Peng and Rui Men and Ruize Gao and Runji Lin and Shijie Wang and Shuai Bai and Sinan Tan and Tianhang Zhu and Tianhao Li and Tianyu Liu and Wenbin Ge and Xiaodong Deng and Xiaohuan Zhou and Xingzhang Ren and Xinyu Zhang and Xipin Wei and Xuancheng Ren and Yang Fan and Yang Yao and Yichang Zhang and Yu Wan and Yunfei Chu and Yuqiong Liu and Zeyu Cui and Zhenru Zhang and Zhihao Fan},
      journal={arXiv preprint arXiv:2407.10671},
      year={2024}
}

@inproceedings{pap5,
  title={Pythia: A suite for analyzing large language models across training and scaling},
  author={Biderman, Stella and Schoelkopf, Hailey and Anthony, Quentin Gregory and Bradley, Herbie and O’Brien, Kyle and Hallahan, Eric and Khan, Mohammad Aflah and Purohit, Shivanshu and Prashanth, USVSN Sai and Raff, Edward and others},
  booktitle={International Conference on Machine Learning},
  pages={2397--2430},
  year={2023},
  organization={PMLR}
}

@article{pap4,
  title={Opt: Open pre-trained transformer language models},
  author={Zhang, Susan and Roller, Stephen and Goyal, Naman and Artetxe, Mikel and Chen, Moya and Chen, Shuohui and Dewan, Christopher and Diab, Mona and Li, Xian and Lin, Xi Victoria and others},
  journal={arXiv preprint arXiv:2205.01068},
  year={2022}
}

@article{pap3,
  title={Textbooks are all you need ii: phi-1.5 technical report},
  author={Li, Yuanzhi and Bubeck, S{\'e}bastien and Eldan, Ronen and Del Giorno, Allie and Gunasekar, Suriya and Lee, Yin Tat},
  journal={arXiv preprint arXiv:2309.05463},
  year={2023}
}

@article{Langlais2025EvenSR,
  title={Even Small Reasoners Should Quote Their Sources: Introducing the Pleias-RAG Model Family},
  author={Pierre-Carl Langlais and Pavel Chizhov and Mattia Nee and Carlos Rosas Hinostroza and Matthieu Delsart and Ir{\`e}ne Girard and Othman Hicheur and Anastasia Stasenko and Ivan P. Yamshchikov},
  journal={ArXiv},
  year={2025},
  volume={abs/2504.18225},
  url={https://api.semanticscholar.org/CorpusID:278129318}
}

@misc{sorstkins2025assessingraghyde1b,
      title={Assessing RAG and HyDE on 1B vs. 4B-Parameter Gemma LLMs for Personal Assistants Integretion}, 
      author={Andrejs Sorstkins},
      year={2025},
      eprint={2506.21568},
      archivePrefix={arXiv},
      primaryClass={cs.CL},
      url={https://arxiv.org/abs/2506.21568}, 
}

@misc{zhang2025jasperstelladistillationsota,
      title={Jasper and Stella: distillation of SOTA embedding models}, 
      author={Dun Zhang and Jiacheng Li and Ziyang Zeng and Fulong Wang},
      year={2025},
      eprint={2412.19048},
      archivePrefix={arXiv},
      primaryClass={cs.IR},
      url={https://arxiv.org/abs/2412.19048}, 
}

@misc{chen2025dragdistillingragslms,
      title={DRAG: Distilling RAG for SLMs from LLMs to Transfer Knowledge and Mitigate Hallucination via Evidence and Graph-based Distillation}, 
      author={Jennifer Chen and Aidar Myrzakhan and Yaxin Luo and Hassaan Muhammad Khan and Sondos Mahmoud Bsharat and Zhiqiang Shen},
      year={2025},
      eprint={2506.01954},
      archivePrefix={arXiv},
      primaryClass={cs.CL},
      url={https://arxiv.org/abs/2506.01954}, 
}

@inproceedings{goddard-etal-2024-arcees,
    title = "Arcee{'}s {M}erge{K}it: A Toolkit for Merging Large Language Models",
    author = "Goddard, Charles  and
      Siriwardhana, Shamane  and
      Ehghaghi, Malikeh  and
      Meyers, Luke  and
      Karpukhin, Vladimir  and
      Benedict, Brian  and
      McQuade, Mark  and
      Solawetz, Jacob",
    editor = "Dernoncourt, Franck  and
      Preo{\c{t}}iuc-Pietro, Daniel  and
      Shimorina, Anastasia",
    booktitle = "Proceedings of the 2024 Conference on Empirical Methods in Natural Language Processing: Industry Track",
    month = nov,
    year = "2024",
    address = "Miami, Florida, US",
    publisher = "Association for Computational Linguistics",
    url = "https://aclanthology.org/2024.emnlp-industry.36",
    doi = "10.18653/v1/2024.emnlp-industry.36",
    pages = "477--485",
}

@inproceedings{muennighoff2023mteb,
  title={Mteb: Massive text embedding benchmark},
  author={Muennighoff, Niklas and Tazi, Nouamane and Magne, Lo{\"\i}c and Reimers, Nils},
  booktitle={Proceedings of the 17th Conference of the European Chapter of the Association for Computational Linguistics},
  pages={2014--2037},
  year={2023}
}

@article{hu2022lora,
  title={Lora: Low-rank adaptation of large language models.},
  author={Hu, Edward J and Shen, Yelong and Wallis, Phillip and Allen-Zhu, Zeyuan and Li, Yuanzhi and Wang, Shean and Wang, Lu and Chen, Weizhu and others},
  journal={ICLR},
  volume={1},
  number={2},
  pages={3},
  year={2022}
}

@article{wei2022emergent,
  title={Emergent abilities of large language models},
  author={Wei, Jason and Tay, Yi and Bommasani, Rishi and Raffel, Colin and Zoph, Barret and Borgeaud, Sebastian and Yogatama, Dani and Bosma, Maarten and Zhou, Denny and Metzler, Donald and others},
  journal={arXiv preprint arXiv:2206.07682},
  year={2022}
}

@article{lewis2020retrieval,
  title={Retrieval-augmented generation for knowledge-intensive nlp tasks},
  author={Lewis, Patrick and Perez, Ethan and Piktus, Aleksandra and Petroni, Fabio and Karpukhin, Vladimir and Goyal, Naman and K{\"u}ttler, Heinrich and Lewis, Mike and Yih, Wen-tau and Rockt{\"a}schel, Tim and others},
  journal={Advances in neural information processing systems},
  volume={33},
  pages={9459--9474},
  year={2020}
}

@inproceedings{rageval,
  title={Rageval: Scenario specific rag evaluation dataset generation framework},
  author={Zhu, Kunlun and Luo, Yifan and Xu, Dingling and Yan, Yukun and Liu, Zhenghao and Yu, Shi and Wang, Ruobing and Wang, Shuo and Li, Yishan and Zhang, Nan and others},
  booktitle={Proceedings of the 63rd Annual Meeting of the Association for Computational Linguistics (Volume 1: Long Papers)},
  pages={8520--8544},
  year={2025}
}

@article{soffer2024scalable,
  title={A scalable framework for benchmarking embedding models for semantic medical tasks},
  author={Soffer, Shelly and Glicksberg, Benjamin S and Kovatch, Patricia and Efros, Orly and Freeman, Robert and Charney, Alexander W and Nadkarni, Girish N and Klang, Eyal},
  journal={medRxiv},
  pages={2024--08},
  year={2024},
  publisher={Cold Spring Harbor Laboratory Press}
}

@article{popqa ,
  title={When Not to Trust Language Models: Investigating Effectiveness and Limitations of Parametric and Non-Parametric Memories },
  author={ Mallen, Alex and Asai,Akari and  Zhong, Victor and Das, Rajarshi and Hajishirzi, Hannaneh and Khashabi, Daniel},
  journal={ arXiv preprint },
  year={ 2022 }
}

@inproceedings{fever,
    author = {Thorne, James and Vlachos, Andreas and Christodoulopoulos, Christos and Mittal, Arpit},
    title = {{FEVER}: a Large-scale Dataset for Fact Extraction and {VERification}},
    booktitle = {NAACL-HLT},
    year = {2018}
}

@article{mergekit,                    
      title={Model Stock: All we need is just a few fine-tuned models}, 
      author={Wang, Cheng and Fang, Xinyu and others},          
      journal={arXiv preprint arXiv:2403.19522},
      year={2024}        }

@inproceedings{pubmedqa,
    title = "{P}ub{M}ed{QA}: A Dataset for Biomedical Research Question Answering",
    author = "Jin, Qiao  and
      Dhingra, Bhuwan  and
      Liu, Zhengping  and
      Cohen, William  and
      Lu, Xinghua",
    editor = "Inui, Kentaro  and
      Jiang, Jing  and
      Ng, Vincent  and
      Wan, Xiaojun",
    booktitle = "Proceedings of the 2019 Conference on Empirical Methods in Natural Language Processing and the 9th International Joint Conference on Natural Language Processing (EMNLP-IJCNLP)",
    month = nov,
    year = "2019",
    address = "Hong Kong, China",
    publisher = "Association for Computational Linguistics",
    url = "https://aclanthology.org/D19-1259/",
    doi = "10.18653/v1/D19-1259",
    pages = "2567--2577"
}

@article{arc,
      author    = {Peter Clark  and Isaac Cowhey and Oren Etzioni and Tushar Khot and
                    Ashish Sabharwal and Carissa Schoenick and Oyvind Tafjord},
      title     = {Think you have Solved Question Answering? Try ARC, the AI2 Reasoning Challenge},
      journal   = {arXiv:1803.05457v1},
      year      = {2018},
}

@inproceedings{asqa,
  title={ASQA: Factoid questions meet long-form answers},
  author={Stelmakh, Ivan and Luan, Yi and Dhingra, Bhuwan and Chang, Ming-Wei},
  booktitle={Proceedings of the 2022 Conference on Empirical Methods in Natural Language Processing},
  pages={8273--8288},
  year={2022}
}

@inproceedings{wikimultihopqa,
    title = "Constructing A Multi-hop {QA} Dataset for Comprehensive Evaluation of Reasoning Steps",
    author = "Ho, Xanh  and
      Duong Nguyen, Anh-Khoa  and
      Sugawara, Saku  and
      Aizawa, Akiko",
    booktitle = "Proceedings of the 28th International Conference on Computational Linguistics",
    month = dec,
    year = "2020",
    address = "Barcelona, Spain (Online)",
    publisher = "International Committee on Computational Linguistics",
    url = "https://www.aclweb.org/anthology/2020.coling-main.580",
    pages = "6609--6625",
}

@article{nq,
    title = "Natural Questions: A Benchmark for Question Answering Research",
    author = "Kwiatkowski, Tom  and
      Palomaki, Jennimaria  and
      Redfield, Olivia  and
      Collins, Michael  and
      Parikh, Ankur  and
      Alberti, Chris  and
      Epstein, Danielle  and
      Polosukhin, Illia  and
      Devlin, Jacob  and
      Lee, Kenton  and
      Toutanova, Kristina  and
      Jones, Llion  and
      Kelcey, Matthew  and
      Chang, Ming-Wei  and
      Dai, Andrew M.  and
      Uszkoreit, Jakob  and
      Le, Quoc  and
      Petrov, Slav",
    editor = "Lee, Lillian  and
      Johnson, Mark  and
      Roark, Brian  and
      Nenkova, Ani",
    journal = "Transactions of the Association for Computational Linguistics",
    volume = "7",
    year = "2019",
    address = "Cambridge, MA",
    publisher = "MIT Press",
    url = "https://aclanthology.org/Q19-1026/",
    doi = "10.1162/tacl_a_00276",
    pages = "452--466"
}

@article{triviaqa,
       author = {{Joshi}, Mandar and {Choi}, Eunsol and {Weld},
                 Daniel and {Zettlemoyer}, Luke},
        title = "{triviaqa: A Large Scale Distantly Supervised Challenge Dataset for Reading Comprehension}",
      journal = {arXiv e-prints},
         year = 2017,
          eid = {arXiv:1705.03551},
        pages = {arXiv:1705.03551},
archivePrefix = {arXiv},
       eprint = {1705.03551},
}

@article{conway2025syftr,
  title={syftr: Pareto-Optimal Generative AI},
  author={Conway, Alexander and Dey, Debadeepta and Hackmann, Stefan and Hausknecht, Matthew and Schmidt, Michael and Steadman, Mark and Volynets, Nick},
  journal={arXiv preprint arXiv:2505.20266},
  year={2025}
}

@article{kaist,
  title={The CoT Collection: Improving Zero-shot and Few-shot Learning of Language Models via Chain-of-Thought Fine-Tuning},
  author={Kim, Seungone and Joo, Se June and Kim, Doyoung and Jang, Joel and Ye, Seonghyeon and Shin, Jamin and Seo, Minjoon},
  journal={arXiv preprint arXiv:2305.14045},
  year={2023}
}

@article{qu2026interactive,
  title={Interactive taxonomy development with hybrid methods},
  author={Qu, Jiaming and Gopinathan, Madhu and Akbar, Shayan Ali and Alonso, Omar},
  year={2026}
}

@article{andonian2023gpt,
  title={GPT-NeoX: Large scale autoregressive language modeling in PyTorch},
  author={Andonian, Alex and Biderman, Stella and Black, Sid and Gali, Preetham and Gao, Leo and Hallahan, Eric and Levy-Kramer, Josh and Leahy, Connor and Nestler, Lucas and Parker, Kip and others},
  journal={Zenodo},
  year={2023}
}

@article{llama3,
  title={The llama 3 herd of models},
  author={Grattafiori, Aaron and Dubey, Abhimanyu and Jauhri, Abhinav and Pandey, Abhinav and Kadian, Abhishek and Al-Dahle, Ahmad and Letman, Aiesha and Mathur, Akhil and Schelten, Alan and Vaughan, Alex and others},
  journal={arXiv preprint arXiv:2407.21783},
  year={2024}
}

@article{shao2024deepseekmath,
  title={Deepseekmath: Pushing the limits of mathematical reasoning in open language models},
  author={Shao, Zhihong and Wang, Peiyi and Zhu, Qihao and Xu, Runxin and Song, Junxiao and Bi, Xiao and Zhang, Haowei and Zhang, Mingchuan and Li, YK and Wu, Yang and others},
  journal={arXiv preprint arXiv:2402.03300},
  year={2024}
}

@article{wang2024model,
  title={Model Stock: All we need is just a few fine-tuned models},
  author={Wang, Dong-Hwan and Isber, Sima and Kim, Jaeho and Moon, Minkyu and Han, Dennis and Choi, Joonmo and Kim, Changick},
  journal={arXiv preprint arXiv:2403.19522},
  year={2024}
}
\bibliographystyle{colm2026_conference}

\appendix
\onecolumn

\appendix

\section*{Appendix}

\section{Dataset Details}
\label{app:dataset_details}

\subsection{Source Datasets}

The simpleCoT dataset aggregates seven diverse source datasets totaling 2,214,941 examples across multiple reasoning domains, including        
  multi-hop question answering (HotpotQA Distractor and FullWiki variants), general instruction following (Alpaca cleaned), mathematical         
  reasoning (Math QA), commonsense narrative understanding (Cosmos QA), diverse instruction patterns (GPTeacher), and evolved complex            
  instructions (Wizard Evol V2). The data processing pipeline consists of three key stages: First, all training splits are loaded directly from  
  their respective Hugging Face repositories using the datasets library. Second, format unification is performed by casting all datasets to a    
  standardized Kaist \cite{kaist} schema comprising three fields: context (system instructions and factual passages), question (user query embedded within     
  context), and ground truth (rationale combined with final answer) ensuring consistent structure across heterogeneous sources. Third, the       
  unified datasets are concatenated, shuffled with a fixed random seed (42) for reproducibility, and split into training and testing subsets     
  using an 80/20 ratio, yielding 1,771,953 training examples and 442,988 test examples. This systematic approach preserves the diversity of      
  reasoning types while enabling standardized model training and evaluation across the entire collection. 

\begin{table}[h]
\centering
\small
\begin{tabular}{lrr}
\toprule
\textbf{Dataset} & \textbf{Samples} & \textbf{\% of Total} \\
\midrule
HotpotQA (Distractor + FullWiki) & 180K & 8.1\% \\
Wizard Evol Instruct V2 & 196K & 8.8\% \\
Alpaca (cleaned) & 52K & 2.3\% \\
Math QA & 37K & 1.7\% \\
GPTeacher & 30K & 1.4\% \\
Cosmos QA & 25K & 1.1\% \\
Kaist (base format) & 1,695K & 76.6\% \\
\midrule
\textbf{Total} & \textbf{2,214,941} & \textbf{100.0\%} \\
\bottomrule
\end{tabular}
\caption{Composition of the \texttt{simpleCoT} training dataset. Kaist provides the base format for unification.}
\label{tab:dataset_composition}
\end{table}

\begin{table}[h]
\centering
\small
\begin{tabular}{lrrr}
\toprule
\textbf{Source} & \textbf{Samples} & \textbf{\% of Total}  \\
\midrule
Wizard Evol Instruct V2 & 196,000 & 8.8\%  \\
HotpotQA (combined) & 180,000 & 8.1\% \\
Alpaca (cleaned) & 52,000 & 2.3\% \\
Math QA & 37,000 & 1.7\%  \\
GPTeacher & 30,000 & 1.4\%  \\
Cosmos QA & 25,000 & 1.1\%  \\
Kaist (diverse) & 1,694,941 & 76.6\%  \\
\bottomrule
\end{tabular}
\caption{Training data distribution across source datasets. Kaist contains 139 diverse sub-tasks.}
\label{tab:source_distribution_app}
\end{table}

\begin{table}[h!]
\centering
\begin{tabular}{lr}
\toprule
\textbf{Metric} & \textbf{Value} \\
\midrule
Total Samples & 2,214,941 \\
Source Datasets & 7 \\
Training Split & 1,771,953 (80\%) \\
Test Split & 442,988 (20\%) \\
Generic Attribution & 36.9\% \\
Specific Citation & 0.5\% \\
No Attribution & 63.1\% \\
Fine-tuning Tokens & 723M \\
Shuffle Seed & 42 \\
\bottomrule
\end{tabular}
\caption{Summary statistics for \texttt{simpleCoT} collection.}
\label{tab:dataset_summary_app}
\end{table}

\subsection{Standard Benchmark Results}
\label{app:data_scale_benchmarks}

\textbf{Benchmarks:}
\begin{itemize}
    \item \textbf{PopQA} (\cite{popqa}): Popular entity questions 
    \item \textbf{TriviaQA} (\cite{triviaqa}): Trivia-style factual questions 
    \item \textbf{Natural Questions (NQ)} (\cite{nq}): Google search queries 
    \item \textbf{WikiMultihopQA} (\cite{wikimultihopqa}): Multi-hop Wikipedia reasoning 
    \item \textbf{FEVER} (\cite{fever}): Fact extraction and verification 
    \item \textbf{ASQA} (4.4K) (\cite{asqa}): Ambiguous questions with long-form answers
    \item \textbf{ARC-Challenge} (\cite{arc}): Science questions (AI2 Reasoning Challenge) 
    \item \textbf{PubMedQA} (\cite{pubmedqa}): Biomedical yes/no/maybe questions (500 samples) 
\end{itemize}

\begin{table}[h!]
\centering
\small
\begin{tabular}{p{1.7cm}p{0.6cm}p{0.8cm}p{0.8cm}p{0.8cm}p{0.8cm}p{0.8cm}p{0.8cm}p{0.8cm}p{0.8cm}p{0.8cm}}
\toprule
\textbf{Model} & \textbf{Data} & \textbf{PopQA} & \textbf{TQA} & \textbf{NQ} & \textbf{ASQA} & \textbf{FEVER} & \textbf{ARC} & \textbf{PMQA} & \textbf{WikiMQA} \\
\midrule
Qwen-0.5B & 50K & 7.90 & 8.60 & 4.20 & 6.00 & 20.53 & \textbf{74.80} & 7.40 & 6.01 \\
Qwen-0.5B GRPO & 50K & 8.82 & 11.60 & 4.20 & 5.40 & 7.39 & 72.00 & 10.30 & 7.06 \\
\midrule
Llama-1B & 50K & 9.60 & 25.00 & 11.60 & 13.40 & 14.26 & 52.60 & 31.40 & 9.56 \\
Llama-1B GRPO & 50K & 11.20 & 25.00 & 11.60 & 13.40 & 15.40 & 52.60 & 32.60 & 13.20 \\
\bottomrule
\end{tabular}
\caption{Standard benchmark results (Exact Match \%) for smaller models trained on 50K subset. GRPO produces modest gains on most tasks but drops on FEVER and ARC for Qwen-0.5B, consistent with the reward not penalizing format deviations.}
\label{tab:data_scale_benchmarks}
\end{table}

\subsection{Note on low/zero scores}
\label{sec:zeroscores}         
  A striking finding across our RAG benchmarks was the widespread failure of small language models to adhere to basic output format constraints, 
  even with extensive prompt engineering. This phenomenon manifested in several distinct failure modes that fundamentally undermined evaluation  
  validity. Most notably, we observed systematic "option repetition" behavior where models would output all possible answer choices rather than  
  selecting one which is a trivial strategy that artificially inflates performance metrics. For instance, on PubMedQA, Llama 3.2 1B repeated all three    
  options ("yes," "no," "maybe") in 321 of 500 test cases, creating false positive matches that initially suggested 80\% accuracy. Only after     
  implementing strict single-answer extraction did we observe the true baseline performance.\\

  \textbf{Catastrophic Failures and Zero-Score Phenomena: }        
  Several model-benchmark combinations produced zero or near-zero scores despite iterative prompt refinement, revealing fundamental limitations  
  in instruction-following capacity for certain small models. On ARC-Challenge, Phi 1.5, Phi 3 Mini, and TinyLlama all achieved 0\% exact match   
  accuracy, while Llama 3.2 1B managed only 0.09\% (1 correct answer out of 1,172 questions). These failures persisted even when simplifying      
  prompts, adding explicit constraints, or restructuring inputs as multiple-choice questions.     
  Particularly illustrative is the case of Pythia 1.4B on PubMedQA, where the model frequently generated only a period (".") with no mention of  
  the required answer options. When constrained through template-based generation, the model instead repeated instructions, specifically 
  outputting "choose between the following three options and only print the correct answer from 'yes'" in 276 of 500 cases. This repetition      
  artificially boosted accuracy to 44\%, which collapsed to 9\% when these echoed instructions were removed from evaluation.                       
   
  OPT 1.3B exhibited similar catastrophic behavior across multiple benchmarks, achieving 0\% on WikiMultihopQA despite various prompt             
  formulations. On FEVER, OPT repeated the entire context verbatim for every test instance under the original prompt, necessitating a complete   
  reformulation. Even after simplification, the model frequently output the system prompt itself as the answer. \\              
  
  \textbf{Model-Specific Failure Modes and Prompt Sensitivity:}    
  Each model exhibited idiosyncratic responses to prompt engineering, requiring bespoke optimization strategies that often contradicted general  
  best practices. Llama 3.2 1B required verbose, markdown-formatted instructions to trigger chain-of-thought reasoning on FEVER, but this same   
  verbosity caused context repetition in other models. Interestingly, while chain-of-thought generation improved Llama's answer quality in some cases, it frequently failed to reach a conclusion producing multi-step reasoning that never selected between "SUPPORTS" or "REFUTES."          
  Additionally, Llama often formatted answers using LaTeX notation (e.g., "The final answer is: $\boxed{SUPPORTS}$"), complicating answer        
  extraction. In the most problematic cases, Llama outputs contained both "SUPPORTS" and "REFUTES" within extended self-contradictory reasoning; 
  penalizing these double-matches reduced FEVER accuracy from 49\% to 23\%.                             
  Qwen 2.5 1.5B demonstrated extreme sensitivity to prompt format. Markdown-structured prompts triggered full context repetition, while simpler  
  templates produced refusal-like outputs that restated the claim in negative form (e.g., "Refutes: The Ukrainian Soviet Socialist Republic was  
  not a founding participant of the UN. The Soviet Union was a founding participant of the UN."). Most surprisingly, using Qwen's documented     
  message format with role tags degraded performance, causing the model to output repetitive Unicode characters ("ți ți ți..."). Only after      
  converting FEVER to explicit multiple-choice format did Qwen produce evaluable outputs.                    
  Phi 1.5 exhibited a distinct hallucination pattern: when presented with two multiple-choice options (A and B), it systematically "completed"   
  the list to four options (adding hallucinated C and D choices) before selecting an answer. For example:    
\begin{verbatim}
  Choose only one option -            
  A. 'SUPPORTS'   
  B. 'REFUTES'.                                         
  Result: C. 'NEITHER'.                  
          D. 'MAYBE'.                                 
          Answer:                                     
          A. 'SUPPORTS'                              
\end{verbatim}

  This behavior increased from 15.98\% accuracy (when evaluating the full response) to 24.88\% when restricting matches to text following          
  "Answer:", suggesting the model reliably answered correctly after completing its hallucinated option set.                                    
   
\textbf{Evaluation Methodology Implications:}                       
  These findings necessitated substantial refinements to evaluation methodology. Initial exact-match scoring proved inadequate, as it could be   
  trivially gamed through option repetition or context echoing. We implemented several countermeasures:                                       
  \begin{enumerate}
  \item Positional answer extraction: Restricting match detection to text following explicit markers like "Answer:" reduced false positives from    
  hallucinated reasoning chains.                                                                                                                 
  \item Penalty for ambiguity: Responses containing multiple contradictory answers (e.g., both "yes" and "no") were marked incorrect, even if one   
  match was correct.                                                                                                                             
  \item Soft matching: For datasets like WikiMultihopQA, we added fuzzy matching to account for semantically equivalent but lexically different answers, revealing substantial performance gaps between strict (7.56\% for B1ade) and soft matching (13.68\% F1).                                
  
  \item Context truncation strategies: CUDA OOM errors for longer contexts (e.g., FEVER Wikipedia articles on OPT) required systematic truncation,  
  introducing an uncontrolled variable that may have degraded performance for context-dependent questions.                                       

  \end{enumerate}

  The most concerning implication is that standard benchmark scores may systematically overestimate small model capabilities when evaluation     
  doesn't account for these failure modes. The PubMedQA case where naive evaluation suggested 80\% accuracy that dropped to actual baseline       
  performance after proper answer extraction demonstrates how easily benchmarks can be inadvertently "gamed" by models that exploit evaluation   
  weaknesses rather than demonstrating genuine comprehension.                                                                                    
       
  \textbf{Performance Variability and Model Selection:}                                                              
  Despite uniform prompt engineering efforts, model performance varied dramatically across benchmarks in ways that defy simple model-size        
  predictions. On ARC-Challenge, Qwen 2.5 1.5B achieved 71.93\% accuracy while larger models like OPT 1.3B achieved only 22.95\%. Similarly, on    
  PopQA, Pythia 1.4B (73.44\%) substantially outperformed the smaller TinyLlama (64.32\%), but this relationship inverted on WikiMultihopQA where  
  TinyLlama's exact match score (0.23\%), while abysmal, exceeded Pythia's 0.06\%.                                                                        
  These inconsistencies suggest that benchmark performance for small models depends more on alignment between model training distribution and    
  task format than on general reasoning capacity. Qwen's strong ARC performance likely reflects exposure to similar multiple-choice formats      
  during training, while its failure on FEVER (requiring the messages format) indicates brittleness when presented with unexpected input structures.\\

  \textbf{Broader Implications for RAG Systems:}                                                                           
  The systematic instruction-following failures documented here pose significant challenges for deploying small models in retrieval-augmented    
  generation systems. The zero-score phenomenon suggests that below a certain capability threshold, prompt engineering alone cannot induce       
  reliable structured output generation. For production RAG systems requiring predictable output formats (e.g., JSON for API responses), these   
  models would require either constrained decoding or downstream post-processing both adding latency and complexity that partially negate the    
  efficiency advantages of smaller models.                                                                  
  Moreover, the model-specific prompt optimization required for even basic functionality undermines the modularity of RAG architectures. A system
   designed around Llama 3.2 1B's verbose markdown preferences would fail catastrophically if swapped to Qwen without prompt reengineering. This 
  brittleness suggests that small model RAG systems may require maintaining model-specific prompt templates and evaluation pipelines, increasing 
  maintenance burden.

\section{B1ade-embed Additional Details}
\label{app:embed_details}

\subsection{MTEB rankings and performance at tasks}

In the current \textsc{MTEB} dashboard, \texttt{B1ade-embed} achieves the following rankings:

\begin{itemize}
\setlength{\itemsep}{2pt}
\setlength{\topsep}{2pt}
\setlength{\parskip}{0pt}
\setlength{\parsep}{0pt}
    \item \#4 overall (Borda ranking)
    \item \#1 on Retrieval overall
    \item \#4 on Pair Classification and Classification
    \item \#3 on \textsc{AskUbuntuDupQuestions} and \textsc{MTOPDomainClassification}
    \item \#2 on \textsc{ArXivHierarchicalClusteringS2S}, \textsc{Arguana}, and \textsc{SprintDuplicateQuestions}
    \item \#1 on \textsc{BIOSSES}, \textsc{Banking77Classification}, \textsc{CQADupstackGamingRetrieval}, \textsc{MindSmallReranking}, \textsc{SciDocs}
    \item High performing on several STS tasks (mean: 78.99, STS: 87.07)
\end{itemize}

\begin{table}[h]
\centering
\small
\begin{tabular}{lcccccccc}
\toprule
\textbf{Model} & \textbf{Avg} & \textbf{Class} & \textbf{Cluster} & \textbf{Pair} & \textbf{Rerank} & \textbf{Retrieval} & \textbf{STS} \\
\midrule
Stella-400M-v5 & \textbf{70.11} & \textbf{86.67} & \textbf{56.70} & \textbf{87.74} & \textbf{60.16} & \textbf{58.97} & 84.22 \\
learning2\_model & \textit{65.39} & 77.75 & 47.96 & 84.53 & 58.50 & \textit{57.91} & 81.43 \\
gte-large-en-v1.5 & \textit{65.39} & 77.75 & 47.96 & 84.53 & 58.50 & \textit{57.91} & 81.43 \\
cde-small-v1 & 65.00 & \textit{81.71} & \textit{48.32} & 84.69 & 56.75 & 53.27 & 81.63 \\
mxbai-embed-large-v1 & 64.68 & 75.64 & 46.71 & 87.20 & \textit{60.11} & 54.39 & \textit{85.00} \\
UAE-Large-V1 & 64.64 & 75.58 & 46.73 & \textit{87.25} & 59.88 & 54.66 & 84.54 \\
bge-large-en-v1.5 & 64.23 & 75.97 & 46.08 & 87.12 & 60.03 & 54.29 & 83.11 \\
\midrule
\textbf{B1ade-embed (ours)} & 64.21 & 75.16 & 46.46 & 87.07 & 60.00 & 53.30 & \textbf{85.04} \\
\bottomrule
\end{tabular}
\caption{B1ade-embed MTEB leaderboard results (56 datasets, all tasks). At the time of writing, B1ade-embed ranked \#9 overall among 400M-parameter models, competitive with leading retrieval models despite zero-training design via parameter fusion. Bold indicates best performance per column. Scores on Classification (12 datasets), Clustering (11), PairClassification (3), Reranking (4), Retrieval (15), and STS (10) tasks shown as averages.}
\label{tab:mteb_results_detailed}
\end{table}

Table~\ref{tab:mteb_results_detailed} shows that Stella-400M-v5 ranks first across six of seven MTEB categories, making it the strongest available retriever in the sub-500M class. This is why we selected Stella-400M-v5 as the comparison point for the retriever ablation in Section~\ref{sec:ablation}: it represents the strongest available retrieval baseline, and demonstrating that B1ade-embed performs comparably in end-to-end RAG (0.654 vs.\ 0.652 average) against it is a strong validation of our zero-training model merging approach. The only category where B1ade-embed outperforms Stella is STS (85.04 vs.\ 84.22), which is directly relevant to RAG retrieval quality --- measuring semantic similarity between query and passage. This suggests that despite lower overall MTEB scores, B1ade-embed is well-suited for RAG-specific retrieval tasks.

\begin{figure}[h]
    \centering
    \includegraphics[width=0.8\linewidth]{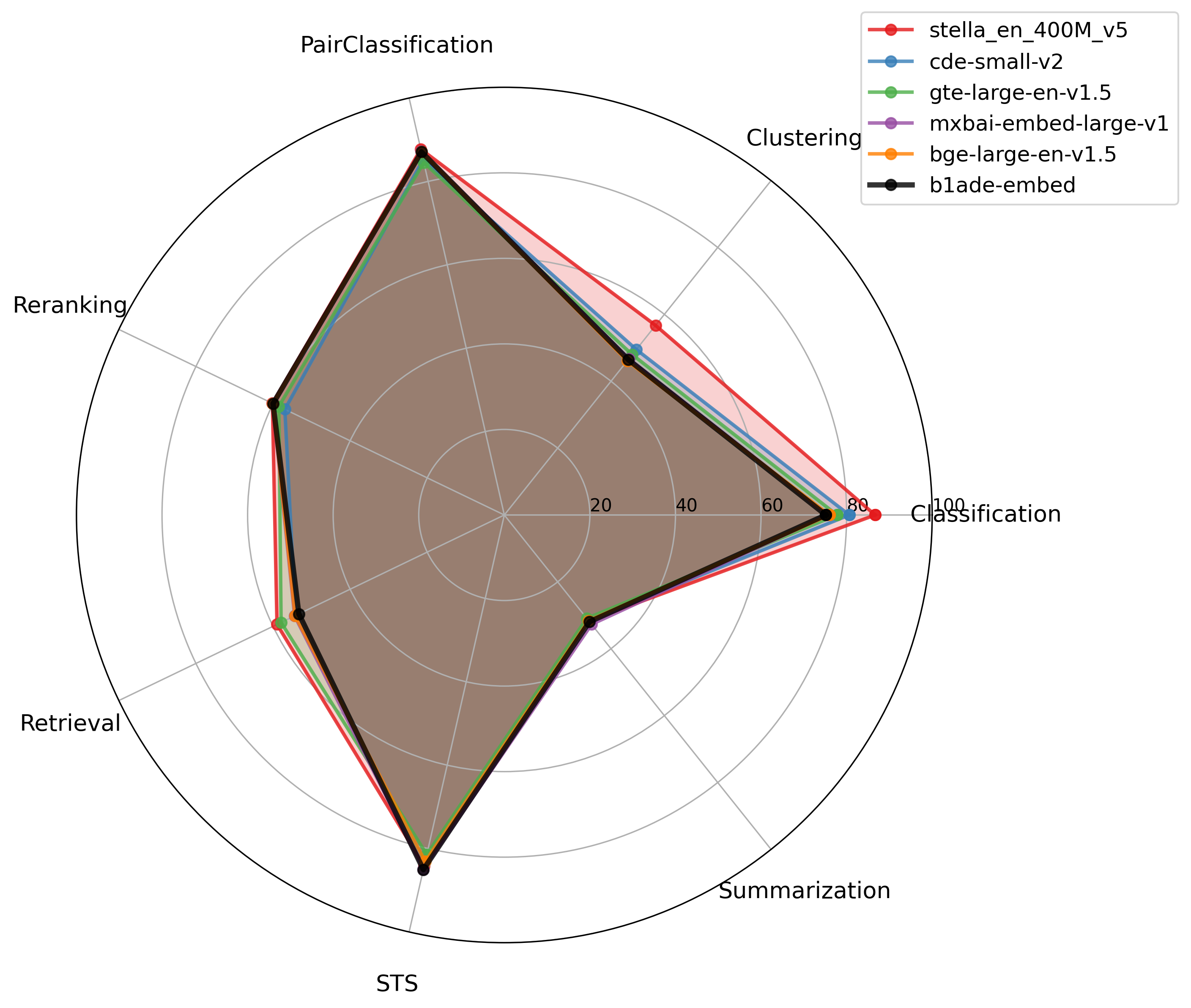}
    \caption{Embedding model performance on legacy MTEB leaderboard for the sub-500M category.}
    \label{fig:modelcomp_app}
\end{figure}

On domain-specific leaderboards (sub-500M class):

\begin{itemize}
\setlength{\itemsep}{2pt}
\setlength{\topsep}{2pt}
\setlength{\parskip}{0pt}
\setlength{\parsep}{0pt}
    \item Top-3 on \textit{CodeSearchNetRetrieval}
    \item Top-5 on legal-text retrieval
    \item Top-2 by mean score on medical benchmarks
    \item \#1 on \textit{NFcorpus}
\end{itemize}

\subsection{Knowledge Distillation Experiments}

We also experimented with distilling knowledge from \texttt{Alibaba-NLP/gte-large-en-v1.5} (434M parameters) into \texttt{b1ade-embed}.        
  Training used AllNLI and English Wikipedia text for 30 epochs (LR = 5e-5, batch size = 64, FP16) with PCA dimensionality reduction followed by 
  dense projection.                                                                                                          
  The distilled model showed targeted gains in specific MTEB subsets, However, average scores dropped across all MTEB categories. All subsequent experiments use the base \texttt{b1ade-embed} model without
  distillation.              

\begin{table}[h]
\centering
\caption{Knowledge distillation results on selected MTEB subsets. The distilled model shows targeted improvements on specific tasks, though average scores dropped across all MTEB categories.}
\label{tab:distillation_results}
\vskip 0.1in
\small
\begin{tabular}{@{}lcccc@{}}
\toprule
\textbf{MTEB Subset} & \textbf{Before (\%)} & \textbf{After (\%)} & \textbf{Improvement} & \textbf{Metric} \\
\midrule
ToxicConversationsClassification & 67.91 & 75.01 & +7.10 & Accuracy \\
TweetSentimentExtractionClassification & 59.22 & 61.8 & +2.58 & Accuracy \\
MedrxivClusteringS2S & 31.7 & 34.02 & +2.32 & Validity Measure \\
ClimateFEVER & 30.37 & 31.95 & +1.58 & nDCG@10 \\
\bottomrule
\end{tabular}
\end{table}

\section{LLMaaJ and Attribution Analysis Examples}
\label{app:attribution_examples}

\subsection{Complete Judge Template}
\label{sec:cmpletetemplate}

We use Claude Sonnet 4 as the judge to rank model outputs. The full prompt template is:

\begin{tcolorbox}[colback=gray!5!white, colframe=gray!75!black, title=Judge Template]
\small
You are evaluating three model responses to a  question. The question includes context passages and asks for a specific answer.

\textbf{Your task:} Rank the three responses from best (Rank 1) to worst (Rank 3) based on these criteria, in order of importance:

\textbf{1. Context Fidelity (HIGHEST PRIORITY)}
\begin{itemize}
    \item Does the response accurately use only the provided context?
    \item Are there any hallucinations or fabricated information?
    \item Does it misquote or misrepresent the passages?
\end{itemize}

\textbf{2. Attribution Quality}
\begin{itemize}
    \item Does the response cite or reference the source passages?
    \item Are claims grounded in specific context?
    \item Note: Prefer responses that acknowledge sources
\end{itemize}

\textbf{3. Completeness}
\begin{itemize}
    \item Does it fully answer the question?
    \item Are all relevant details from context included?
    \item \textbf{DO NOT prefer brevity over completeness}
\end{itemize}

\textbf{4. Factual Accuracy}
\begin{itemize}
    \item Is the core answer correct based on the context?
    \item Does it identify the right entity/relationship/fact?
\end{itemize}

\textbf{Output Format:}
\begin{verbatim}
Model A: Rank [1/2/3]
Model B: Rank [1/2/3]
Model C: Rank [1/2/3]

Reasoning: [Brief explanation of ranking decision]
\end{verbatim}
\end{tcolorbox}

\subsection{Example Evaluations}
\label{app:examples}
\textbf{Example 1: B1ade-1B Wins with Correct Nuance}
\begin{tcolorbox}[colback=blue!5!white, colframe=blue!75!black, title=Context (Excerpt)]
\small
The Asian genotype became the predominant genotype circulating worldwide in 2007...
\end{tcolorbox}

\begin{tcolorbox}[colback=blue!5!white, colframe=blue!75!black, title=Question]
When did the Asian genotype emerge?
\end{tcolorbox}

\textbf{Model Outputs:}

\begin{tcolorbox}[colback=red!5!white, colframe=red!75!black, title=Llama-1B (Rank 3)]
The Asian genotype emerged in Cambodia between 2001 and 2007.
\end{tcolorbox}

\begin{tcolorbox}[colback=orange!5!white, colframe=orange!75!black, title=Qwen-1.5B (Rank 2)]
The Asian genotype emerged in Indochina around 2007.
\end{tcolorbox}

\begin{tcolorbox}[colback=green!5!white, colframe=green!75!black, title=B1ade-1B (Rank 1) $\checkmark$]
Based on the passage, the Asian genotype became the predominant genotype in 2007. However, the passage does not explicitly mention when the Asian genotype first emerged, only when it became predominant.
\end{tcolorbox}

\textbf{Judge Reasoning:}
``B1ade-1B correctly identifies that the passage describes when the genotype became \textit{predominant}, not when it \textit{emerged}. The question asks about emergence, but the context only provides information about predominance. B1ade's careful distinction and epistemic humility (acknowledging the passage doesn't say when it emerged) demonstrate superior context fidelity. Ranks: B1ade=1, Qwen=2, Llama=3.''

\textbf{Analysis:} B1ade wins by understanding the nuance between ``emerged'' and ``became predominant.'' This is the most common winning pattern (48\% of wins).

\textbf{Example 2: B1ade-1B Loses from Factual Error}

\begin{tcolorbox}[colback=blue!5!white, colframe=blue!75!black, title=Context (Excerpt)]
\small
Hemagglutinin (HA) is the major surface and attachment glycoprotein on influenza virus... Neuraminidase (NA) is another surface glycoprotein that facilitates viral release.
\end{tcolorbox}

\begin{tcolorbox}[colback=blue!5!white, colframe=blue!75!black, title=Question]
Which is the major surface and attachment glycoprotein on influenza virus?
\end{tcolorbox}

\textbf{Model Outputs:}

\begin{tcolorbox}[colback=green!5!white, colframe=green!75!black, title=Llama-1B (Rank 1)]
Hemagglutinin (HA)
\end{tcolorbox}

\begin{tcolorbox}[colback=green!5!white, colframe=green!75!black, title=Qwen-1.5B (Rank 1)]
hemagglutinin (HA)
\end{tcolorbox}

\begin{tcolorbox}[colback=red!5!white, colframe=red!75!black, title=B1ade-1B (Rank 3)]
Based on the passage, the NA glycoprotein is the major surface and attachment glycoprotein on influenza virus.
\end{tcolorbox}

\textbf{Judge Reasoning:}
``Context explicitly states: 'Hemagglutinin (HA) is the major surface and attachment glycoprotein.' B1ade-1B's answer of NA (neuraminidase) is factually incorrect. Despite proper attribution (``Based on the passage''), the core answer is wrong. Llama and Qwen both correctly identify HA. Ranks: Llama=1, Qwen=1, B1ade=3.''

\textbf{Analysis:} B1ade-1B loses from factual error (56\% of losses), not from attribution strategy. The model correctly attributes but provides the wrong glycoprotein.

\textbf{Example 3: B1ade-1B Wins Despite Attribution Paradox}

\begin{tcolorbox}[colback=blue!5!white, colframe=blue!75!black, title=Question]
What percentage of patients developed acute liver injury?
\end{tcolorbox}

\textbf{Model Outputs:}

\begin{tcolorbox}[colback=orange!5!white, colframe=orange!75!black, title=Qwen-1.5B (Rank 2)]
14-53\% of patients with COVID-19 developed acute liver injury.
\end{tcolorbox}

\begin{tcolorbox}[colback=green!5!white, colframe=green!75!black, title=B1ade-1B (Rank 1) ]
According to the passage, 14-53\% of patients with COVID-19 developed acute liver injury, with elevated levels of ALT and AST.
\end{tcolorbox}

\textbf{Judge Reasoning:}
``Both provide the correct range (14-53\%). B1ade adds useful clinical detail (elevated ALT/AST) and explicitly attributes to the passage. More complete response. Ranks: B1ade=1, Qwen=2.''

\section{Attribution Analysis:} B1ade-1B wins when attribution is combined with additional relevant detail from context. This shows attribution can provide value when execution is correct.

\subsection{Detection Keywords}

\textbf{Generic Attribution Phrases:}
\begin{itemize}
    \item ``based on'', ``according to'', ``the passage'', ``the context''
    \item ``as mentioned'', ``as stated'', ``from the passage'', ``the article''
    \item ``as described'', ``in the passage'', ``the text states''
\end{itemize}

\textbf{Specific Citation Phrases:}
\begin{itemize}
    \item ``passage 1'', ``passage 2'', ``passage 3''
    \item ``first passage'', ``second passage'', ``third passage''
    \item ``title:'', ``paragraph''
\end{itemize}

\textbf{Detection Method:} Case-insensitive substring matching. A response is classified as ``attributed'' if it contains any generic attribution phrase. If it also contains a specific citation phrase, it's classified as ``specifically cited.''

\subsection{Quote Accuracy Measurement}

We measure quote accuracy using fuzzy string matching:

\begin{lstlisting}[language=Python, basicstyle=\small\ttfamily, frame=single]
from difflib import SequenceMatcher

def compute_quote_accuracy(quote, passages):
    """
    Check if quote appears in any passage.
    Returns accuracy category.
    """
    max_overlap = 0.0

    for passage in passages:
        # Normalize both strings
        quote_norm = normalize(quote)
        passage_norm = normalize(passage)

        # Compute sequence similarity
        similarity = SequenceMatcher(
            None, quote_norm, passage_norm
        ).ratio()

        max_overlap = max(max_overlap, similarity)

    # Classify
    if max_overlap >= 0.8:
        return "exact"      # High overlap
    elif max_overlap >= 0.5:
        return "partial"    # Medium overlap (paraphrase)
    else:
        return "inaccurate" # Low overlap (misquote)
\end{lstlisting}

\textbf{Thresholds:}
\begin{itemize}
    \item $\geq$80\%: Exact or very close quote
    \item 50-80\%: Partial quote or paraphrase
    \item $<$50\%: Inaccurate or fabricated
\end{itemize}

\subsection{Attribution Analysis Details}
\label{app:attribution_detailed}

\subsubsection{Manual Pattern Analysis}
\label{app:manual_analysis}

We manually analyzed 50 cases (25 wins, 25 losses) to identify qualitative patterns.

\textbf{Why B1ade Wins:}

\begin{table}[h!]
\centering
\begin{tabular}{lrp{6cm}}
\toprule
\textbf{Pattern} & \textbf{\% of Wins} & \textbf{Description} \\
\midrule
Correct Nuance & \textbf{48\%} & Understands what question actually asks \\
Complete Answer & 16\% & Provides comprehensive response \\
Least Wrong & 16\% & All models struggle; B1ade closest \\
Direct Quote & 13\% & Includes exact passage quote \\
Good Attribution & 6\% & Clear source citation helps \\
\bottomrule
\end{tabular}
\caption{Patterns in B1ade winning cases.}
\label{tab:win_patterns_app}
\end{table}

\textbf{Why B1ade Loses:}

\begin{table}[h!]
\centering
\begin{tabular}{lrp{6cm}}
\toprule
\textbf{Pattern} & \textbf{\% of Losses} & \textbf{Description} \\
\midrule
Factually Wrong & \textbf{56\%} & Incorrect answer to question \\
Incomplete & 20\% & Missing key information \\
Misunderstood Question & 12\% & Answered different question \\
Over-Interpretation & 12\% & Added unsupported inference \\
\textbf{Fabricated Sources} & \textbf{0\%} & Made up citations \\
\bottomrule
\end{tabular}
\caption{Patterns in B1ade losing cases. Critical finding: B1ade loses from factual errors (56\%), NOT from fabrication (0\%).}
\label{tab:loss_patterns_app}
\end{table}

\subsubsection{Attribution results}
\begin{table}[h]
\centering
\begin{tabular}{lcccc}
\toprule
\textbf{Metric} & \textbf{B1ade} & \textbf{Qwen} & \textbf{Llama} & \textbf{Training} \\
\midrule
\textbf{Attribution Usage} & \cellcolor{green!20}\textbf{42.4\%} & 40.9\% & 35.4\% & 36.9\% \\
Total Attributions & 2,689 & 2,595 & 2,243 & --- \\
\midrule
Good Attributions & 1,825 & \cellcolor{green!20}\textbf{2,126} & 1,624 & --- \\
Bad Attributions & 592 & 227 & 380 & --- \\
\midrule
\textbf{Success Rate} & 67.9\% & \cellcolor{green!20}\textbf{81.9\%} & 72.4\% & --- \\

\bottomrule
\end{tabular}
\caption{Complete attribution analysis across models.}
\label{tab:attribution_complete}
\end{table}

\begin{figure*}[h]
\centering
\includegraphics[width=0.85\textwidth]{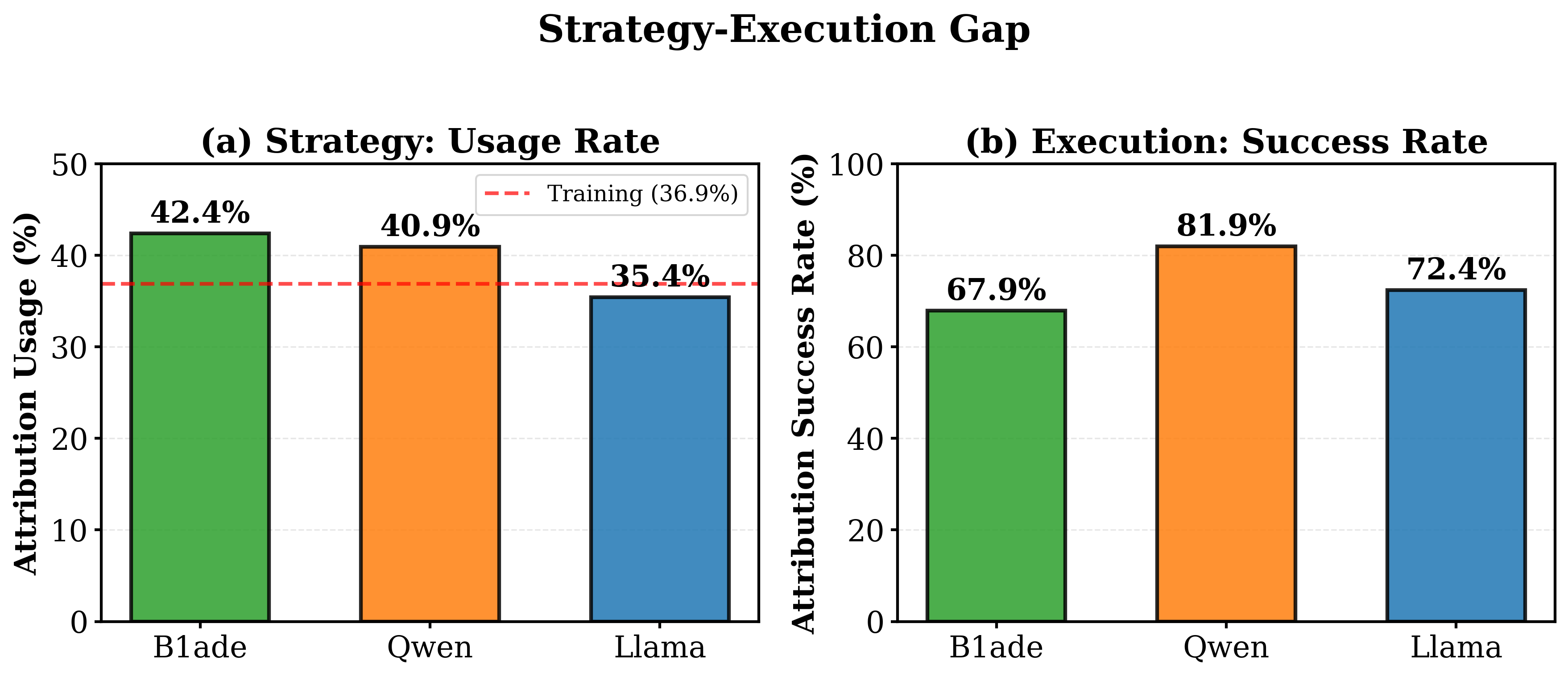}
\caption{Strategy-execution gap in attribution behavior. (a) B1ade demonstrates the highest attribution usage rate (42.4\%), exceeding both the training baseline (36.9\%, red dashed line) and other models, validating superior strategic application. (b) However, Qwen achieves the highest success rate (81.9\%) when attributing, revealing an execution quality gap.}
\label{fig:strategy_execution_app}
\end{figure*}

\begin{table}[h]
\centering
\begin{tabular}{lcccc}
\toprule
\textbf{Metric} & \textbf{B1ade} & \textbf{Qwen} & \textbf{Llama} \\
\midrule
Total Quotes & \cellcolor{green!20}\textbf{1,274} & 851 & 909 \\
Quote Usage \% & \cellcolor{green!20}\textbf{9.8\%} & 7.0\% & 7.4\% \\
\midrule
Accurate Quotes & \cellcolor{green!20}\textbf{494} & 419 & 325 \\
Quote Accuracy \% & 38.8\% & \textbf{49.2\%} & 35.8\% \\
\midrule
Rank Improvement & \cellcolor{green!20}\textbf{+0.105} & +0.098 & -0.184 \\
\bottomrule
\end{tabular}
\caption{Quote usage and effectiveness. B1ade quotes most frequently and has most accurate quotes in absolute terms (494).}
\label{tab:quote_analysis_app}
\end{table}

To illustrate the attribution patterns observed in the training data, we present three representative examples from the ground truth responses.
   These examples demonstrate the types of attribution behaviors models learn to emulate. Example 1 and 3 show explicit source attribution using phrases like ``the article states that'' and ``the context describes,''
  while Example 2 represents the majority case (63.1\%) where responses provide information without explicit attribution markers.

\textbf{Example 1: With Generic Attribution (36.9\%)}

\textit{Ground Truth:} ``\textcolor{blue}{The article states that} many medicinal and recreational drugs, such as tetrahydrocannabinol (active ingredient in cannabis), caffeine, morphine and nicotine come directly from plants. These are some examples of the medicines found in plants mentioned by the author.''

\textbf{Example 2: No Attribution (63.1\%)}

\textit{Ground Truth:} ``The Battle of Appomattox Court House was a battle in the final stages of the American Civil War, resulting in Confederate General Robert E. Lee surrendering his Army to Union Commander Ulysses S. Grant on April 9th 1865.''

\textbf{Example 3: With Context Reference (36.9\%)}

\textit{Ground Truth:} ``\textcolor{blue}{The context describes} events and details of his life, such as how he met Edna Stillwell in Vincennes and their early careers, the `Doughnut Dunkers' routine that brought him recognition...''

\section{Evaluation Bias Analysis}
\label{sec:biasllm}
We identify several systematic biases in judge behavior:

\subsection{Length Bias}

Despite explicit instruction ``DO NOT prefer brevity over completeness,'' we observe correlation between response length and worse ranks for some models. For example Llama's longer responses are penalized despite judge instructions.

\begin{table}[t!]
\centering
\begin{tabular}{lccc}
\toprule
\textbf{Model} & \textbf{Avg Length} & \textbf{Length-Rank Correlation} & \textbf{Bias} \\
\midrule
B1ade & 538 chars & +0.084 & Slight penalty \\
Llama & 500 chars & +0.264 & Strong penalty \\
Qwen & 322 chars & -0.112 & Slight reward \\
\bottomrule
\end{tabular}
\caption{Length bias analysis. Positive correlation = longer responses receive worse ranks.}
\label{tab:length_bias}
\end{table}

\subsection{Attribution Paradox}

We manually found 10+ cases where B1ade explicitly attributed the right excerpt but lost to Qwen without attribution, contradicting the stated preference for source citation.

\textbf{Example Case 1:}

\begin{tcolorbox}[colback=blue!5!white, colframe=blue!75!black, title=Question]
How many antigens could Liew's ELISA detect?
\end{tcolorbox}

\begin{tcolorbox}[colback=green!5!white, colframe=green!75!black, title=Qwen (Rank 1 - No Attribution)]
Liew validated one multiplex ELISA for the detection of 9 antigens.
\end{tcolorbox}

\begin{tcolorbox}[colback=orange!5!white, colframe=orange!75!black, title=B1ade (Rank 2 - With Attribution)]
Based on the passage, Liew's multiplex ELISA test could detect 9 antigens.
\end{tcolorbox}

Both responses are factually identical; B1ade adds attribution phrase. Judge ranks Qwen higher despite stated preference for attribution.

\textbf{Example Case 2:}

\begin{tcolorbox}[colback=blue!5!white, colframe=blue!75!black, title=Question]
It continues to be a challenge for clinicians to identify preterm infants likely to experience subsequent neurodevelopmental deficits. The Test of Infant Motor Performance (TIMP) and the assessment of spontaneous general movements (GMs) are the only reliable diagnostic and predictive tools for the functionality of the developing nervous system, if applied before term.AIM: To determine to what extent singular preterm assessments of motor performance can predict the neurodevelopmental outcome in 14-month olds. The assessment of the quality of general movements (GMs) in young infants is a reliable and valid diagnostic tool for detecting brain dysfunction early in life. Poor repertoire GMs are the most frequently observed abnormal GMs during the preterm, term and early postterm period. However, their predictive value for the neurological outcome is low.AIM: To find out whether a detailed scoring of poor repertoire GMs might lead to a better prediction of the neurological outcome. To assess the clinical value of a modified version, not employing video recording, of Precthl's method on the qualitative assessment of general movements (GMs) in preterm, term and young infants at neurological risk. \textbf{Is it possible to predict the infant's neurodevelopmental outcome at 14 months of age by means of a single preterm assessment of General Movements?}
\end{tcolorbox}

\begin{tcolorbox}[colback=green!5!white, colframe=green!75!black, title=Qwen (Rank 1 - with Attribution)]
Yes, according to Passage \#1, it is possible to predict the infant's neurodevelopmental outcome at 14 months of age using a single preterm assessment of General Movements (GMs).
\end{tcolorbox}

\begin{tcolorbox}[colback=orange!5!white, colframe=orange!75!black, title=B1ade (Rank 2 - With Attribution)]
Based on the provided context and the details given about the Test of Infant Motor Performance (TIMP) and the assessment of spontaneous general movements (GMs), it appears that the current approach is to use a single assessment of general movements to predict neurodevelopmental deficits in preterm infants at 14 months of age. The assessment of the quality of general movements (GMs) in young infants is a reliable and valid diagnostic tool for detecting brain dysfunction early in life, but its predictive value for neurological outcome is low. To address the question of whether it is possible to predict the infant's neurodevelopmental outcome at 14 months of age by means of a single preterm assessment of General Movements, the answer is: No, it is not possible to predict the infant's neurodevelopmental outcome at 14 months of age by means of a single preterm assessment of General Movements.
\end{tcolorbox}

In this case Qwen's answer is in fact wrong, despite pointing to the passage. B1ade correctly answers this nuanced question very similar to the ground truth:

\begin{tcolorbox}[colback=gray!5!white, colframe=gray!75!black, title=Ground truth]
Based on the context provided, it seems that the ability to predict an infant's neurodevelopmental outcome at 14 months of age using a single preterm assessment of General Movements (GMs) is limited. The key points are: 1. The assessment of spontaneous GMs is one of the only reliable diagnostic and predictive tools for the functionality of the developing nervous system in preterm infants, if applied before term. 2. However, the context states that "Poor repertoire GMs are the most frequently observed abnormal GMs during the preterm, term and early postterm period. However, their predictive value for the neurological outcome is low." 3. The first study mentioned aimed to determine to what extent singular preterm assessments of motor performance can predict the neurodevelopmental outcome in 14-month-olds. This suggests that a single preterm GM assessment may have limited predictive value for 14-month neurodevelopmental outcomes. Therefore, based on the information provided, it does not seem possible to reliably predict a preterm infant's neurodevelopmental outcome at 14 months of age using a single preterm assessment of General Movements alone. The predictive value of this assessment appears to be limited, and multiple assessments over time may be needed for more accurate predictions of longer-term neurodevelopmental outcomes.
\end{tcolorbox}


\textbf{Example Case 3:}

\begin{tcolorbox}[colback=blue!5!white, colframe=blue!75!black, title=Question]
Kodiak Island (Alutiiq: "Qikertaq", ) is a large island on the south coast of the U.S. state of Alaska, separated from the Alaska mainland by the Shelikof Strait. The largest island in the Kodiak Archipelago, Kodiak Island is the second largest island in the United States and the 80th largest island in the world, with an area of , slightly larger than Cyprus. It is long and in width ranges from . Kodiak Island is the namesake for Kodiak Seamount, which lies off the coast at the Aleutian Trench. The largest community on the island is the city of Kodiak, Alaska. The U.S. state of Florida has a total of 4,510 islands that are ten acres or larger. This is the second-highest number of islands of any state of the United States; only Alaska has more. Kings Island is a amusement park located northeast of Cincinnati in Mason, Ohio. Owned and operated by Cedar Fair, the park first opened in 1972 by the Taft Broadcasting Company. It was part of a larger effort to move and expand Coney Island, a popular resort destination along the banks of the Ohio River that was prone to frequent flooding. After more than \$300 million in capital investments, the park features over 100 attractions including fourteen roller coasters and a water park. \textbf{How large is King's Island?}
\end{tcolorbox}

\begin{tcolorbox}[colback=green!5!white, colframe=green!75!black, title={Qwen (Rank 3 - hallucination, overconfidence)}]
King's Island is approximately 100 acres in size.
\end{tcolorbox}

\begin{tcolorbox}[colback=orange!5!white, colframe=orange!75!black, title=B1ade (Rank 2 - brief answer with attribution)]
The answer to the question "How large is King's Island?" is not explicitly mentioned in the given text.
\end{tcolorbox}

In this case Qwen's answer is hallucinated. B1ade correctly and briefly answers this question. Here the ground truth is more explanatory but matches B1ade's answer. 

\begin{tcolorbox}[colback=gray!5!white, colframe=gray!75!black, title=Ground truth]
The first context discusses the physical characteristics of the King Island emu, which provides information about the size of the bird's bones, but does not relate to the size of the amusement park. The second context describes King's Island as an amusement park with over 100 attractions, including fourteen roller coasters and a water park. However, it does not provide any information about the physical size of the park.Therefore, there is no available information to provide a direct answer to the question "How large is King's Island?" based on the two given contexts. [No reference available].
\end{tcolorbox}


\section{B1ade-embed: Mergekit Configuration Details}
\label{app:mergekit_details}

\subsection{Model Stock Merging Strategy}

\textit{B1ade-embed} employs the \emph{Model Stock} algorithm~\cite{wang2024model}, which optimizes linear interpolation weights through geometric properties of task vectors. Model Stock provides a principled, data-driven approach to combining multiple pretrained models without requiring additional training, making it ideal for efficient model composition.

\subsubsection{Merging Configuration}

The merge configuration uses Mergekit\footnote{\url{https://github.com/arcee-ai/mergekit}}~\cite{goddard-etal-2024-arcees}, an open-source toolkit for parameter-efficient model composition:

\begin{itemize}
\setlength{\itemsep}{2pt}
\setlength{\parskip}{0pt}
\item \textbf{Merge Method}: \texttt{model\_stock} — geometric weight optimization for linear interpolation
\item \textbf{Base Model}: \texttt{bert-large-uncased} — serves as reference for task vector computation and architecture anchor
\item \textbf{Auxiliary Models}: Four MTEB-optimized dense retrievers:
  \begin{itemize}
  \setlength{\itemsep}{1pt}
    \item \texttt{WhereIsAI/UAE-Large-V1} — specialized in multilingual dense retrieval and cross-language transfer
    \item \texttt{BAAI/bge-large-en-v1.5} — optimized for cross-lingual semantic matching and zero-shot retrieval
    \item \texttt{mixedbread-ai/mxbai-embed-large-v1} — instruction-following embeddings for task-aware representations
    \item \texttt{avsolatorio/GIST-large-Embedding-v0} — general-purpose semantic similarity and broad domain coverage
  \end{itemize}
\item \textbf{Output Dimension}: 1024 (matching BERT-Large architecture)
\item \textbf{Precision}: \texttt{float32} for numerical stability when combining diverse embedding spaces
\item \textbf{Tokenizer}: Unified to \texttt{bert-large-uncased} tokenizer (30,522 vocabulary)
\end{itemize}

\subsubsection{Model Stock Algorithm}

Model Stock computes optimized interpolation weights using the following procedure:

\begin{enumerate}
\item \textbf{Task Vector Extraction}: For each auxiliary model $i$, compute the task vector as the difference from the base model:
\begin{equation}
\tau_i = \theta_i - \theta_{\text{base}}
\end{equation}
where $\theta_i$ and $\theta_{\text{base}}$ are the flattened parameter vectors.

\item \textbf{Pairwise Similarity Computation}: Calculate cosine similarities between all pairs of task vectors to measure model coherence:
\begin{equation}
s_{ij} = \frac{\tau_i \cdot \tau_j}{\|\tau_i\| \cdot \|\tau_j\|}
\end{equation}

\item \textbf{Interpolation Weight Derivation}: The interpolation factor $t$ is computed as the mean of pairwise similarities, which reflects how well-aligned the auxiliary models are:
\begin{equation}
t = \frac{1}{\binom{n}{2}} \sum_{i < j} s_{ij}
\end{equation}

\item \textbf{Weighted Linear Merge}: The final merged parameters are computed as a weighted interpolation between the base model and the average of auxiliary models:
\begin{equation}
\theta_{\text{merged}} = t \cdot \bar{\theta}_{\text{aux}} + (1-t) \cdot \theta_{\text{base}}
\end{equation}
where $\bar{\theta}_{\text{aux}} = \frac{1}{n}\sum_{i=1}^{n} \theta_i$ is the arithmetic mean of auxiliary model parameters.
\end{enumerate}

This adaptive weighting mechanism automatically balances the contribution from all auxiliary models based on their mutual alignment, avoiding manual hyperparameter tuning while maintaining base model coherence throughout the merged architecture.

\subsubsection{Model Selection Rationale}

The five models (including base) were selected based on complementary MTEB performance characteristics:

\begin{itemize}
\setlength{\itemsep}{2pt}
\item \textbf{Diversity}: Each model excels on different MTEB task types (retrieval, similarity, clustering, classification, reranking), ensuring broad semantic coverage
\item \textbf{Efficiency}: All models remain below 1024-dimensional output and compatible with BERT-Large architecture, enabling parameter-free composition
\item \textbf{Performance}: Collectively rank in the top tier of MTEB leaderboard within the sub-500M parameter constraint at the time of the experiment
\item \textbf{Complementarity}: Task vector analysis revealed high diversity (low pairwise similarity) while remaining semantically coherent, indicating each model contributes unique knowledge
\end{itemize}

The resulting 335M-parameter merged model inherits representational capabilities from all five input models without requiring additional pretraining, demonstration, or fine-tuning, making it highly resource-efficient for downstream RAG applications.

\end{document}